\documentclass{article}
\PassOptionsToPackage{authoryear,round}{natbib}
\usepackage[main,final]{style}
\usepackage{xcolor}
\usepackage{float}
\usepackage{multirow}
\usepackage{graphicx}
\usepackage{subcaption}
\usepackage[shortlabels]{enumitem}
\setlist{itemsep=0pt, leftmargin=15pt, topsep=5pt,}
\setlist[1]{labelindent=\parindent}
\usepackage[
  colorlinks=true,
  linkcolor=blue!60!black,
  citecolor=blue!50!black,
  urlcolor=blue!70!black
]{hyperref}

\newcommand{\vsbase}[1]{\textcolor{teal}{\small$\uparrow$#1}}

\usepackage{xcolor}

\newcommand{\loss}[1]{\small\textcolor{orange!80!red}{\small$\downarrow$#1}}

\usepackage[utf8]{inputenc} 
\usepackage[T1]{fontenc}    
\usepackage{hyperref}       
\usepackage{url}            
\usepackage{booktabs}       
\usepackage{amsfonts}       
\usepackage{nicefrac}       
\usepackage{microtype}      
\usepackage{xcolor}         
\usepackage{amsmath}

\title{SCENT: Aligning Mass Spectra with Molecular Structure for Olfactory Perception}

\author{%
  Ziqi Zhang$^1$ \quad Eunyeong Jin$^2$ \quad Miguel Vasco$^1$ \quad
  Farzaneh Taleb$^1$ \quad Nona Rajabi$^1$ \\
  \textbf{Alexandra Gutmann}$^2$ \quad \textbf{Jonathan Williams}$^2$ \quad
  \textbf{Ant\^{o}nio H. Ribeiro}$^{3,4}$ \quad \textbf{Danica Kragic}$^1$ \\[3pt]
  {\small $^1$Dept.\ of Intelligent Systems, KTH Royal Institute of Technology, Stockholm, Sweden} \\
  {\small $^2$Atmospheric Chemistry Dept., Max Planck Institute for Chemistry, Mainz, Germany} \\
  {\small $^3$Dept.\ of Information Technology, Uppsala University, Uppsala, Sweden} \\
  {\small $^4$Science for Life Laboratory (SciLifeLab), Uppsala, Sweden} \\[2pt]
  {\small \texttt{ziqizh@kth.se, miguelsv@kth.se, farzantn@kth.se, nonar@kth.se, dani@kth.se}} \\
  {\small \texttt{eunyeong.jin@mpic.de, a.gutmann@mpic.de, jonathan.williams@mpic.de}} \\
  {\small \texttt{antonio.horta.ribeiro@it.uu.se}}
}

\begin{document}

\maketitle

\begin{abstract}
Predicting human olfactory perception from molecular structure has seen remarkable progress, yet these approaches require explicit chemical structure at inference, which is not available in practical sensing settings. We address this gap by exploring direct electron ionization mass spectrometry (EI-MS), a sensing technique that acquires chemically informative fragmentation fingerprints in seconds, as an alternative input modality for olfactory prediction. We contribute Spectrum-to-Chemical Embedding alignmeNT (SCENT), a multi-modal contrastive learning framework that aligns EI-MS representations with pretrained chemical structure embeddings, while requiring only mass spectra at inference. On the multi-label odor descriptor prediction task, SCENT significantly outperforms MS-only baselines and achieves performance comparable to structure-based models, despite requiring no explicit molecular structure at test time. The learned representations also better approximate continuous human perceptual ratings and generalize to real-world lab-measured spectra, suggesting that cross-modal alignment is an effective strategy for grounding analytical spectra in chemical semantics.
\end{abstract}

\section{Introduction}

Human olfaction remains one of the least explored perceptual modalities in machine learning research, and its computational modeling still faces many open challenges. A widely accepted hypothesis in olfactory neuroscience holds that the perceptual quality of an odor is primarily determined by the physicochemical structure of the odorant molecule ~\citep{rossiter1996structure, suh2025comparative}, which provides the theoretical foundation for the quantitative structure-odor relationship (QSOR) paradigm and has attracted increasing attention by the machine learning community in recent years~\citep{dream,sanchez2019machine, Lee2023,taleb2024can, zhang2024deep}. However, all these approaches assume access to the molecular identity or explicit chemical structure of the odorant at inference time, which demands specialized equipment, chemical expertise, and several hours of acquisition time. This creates a fundamental gap between current structure-based olfactory models and real-world deployable sensing systems.

To address this gap, we contribute, to the best of our knowledge, the first computational olfactory model that leverages high-quality structural information at training time and rapid signal-based measurements at inference to predict odor-related properties. Electronic nose and gas sensor array systems have been widely used for rapid measurements, yet produce non-specific signals with limited chemical interpretability~\citep{yan2015electronic, sung2024data}. Instead, we leverage direct electron ionization mass spectrometry with a quadrupole (direct EI-MS, hereafter MS), which retains chemically relevant information while enabling single-molecule spectra to be acquired in seconds~\citep{demarque2016fragmentation, ji2020predicting}. Existing end-to-end approaches that predict odor labels directly from MS~\citep{nozaki2016odor, ito2020improvement, hasebe2022exploration} rely on autoencoders, treating the spectrum as a raw input signal without explicitly leveraging the implicit chemical information it encodes. This raises a natural question: \emph{can we learn a model that encodes chemical structure into MS representations at training time, such that MS alone is sufficient for odor prediction at inference?}

\begin{figure}[t]
    \centering
    \includegraphics[width=0.95\linewidth]{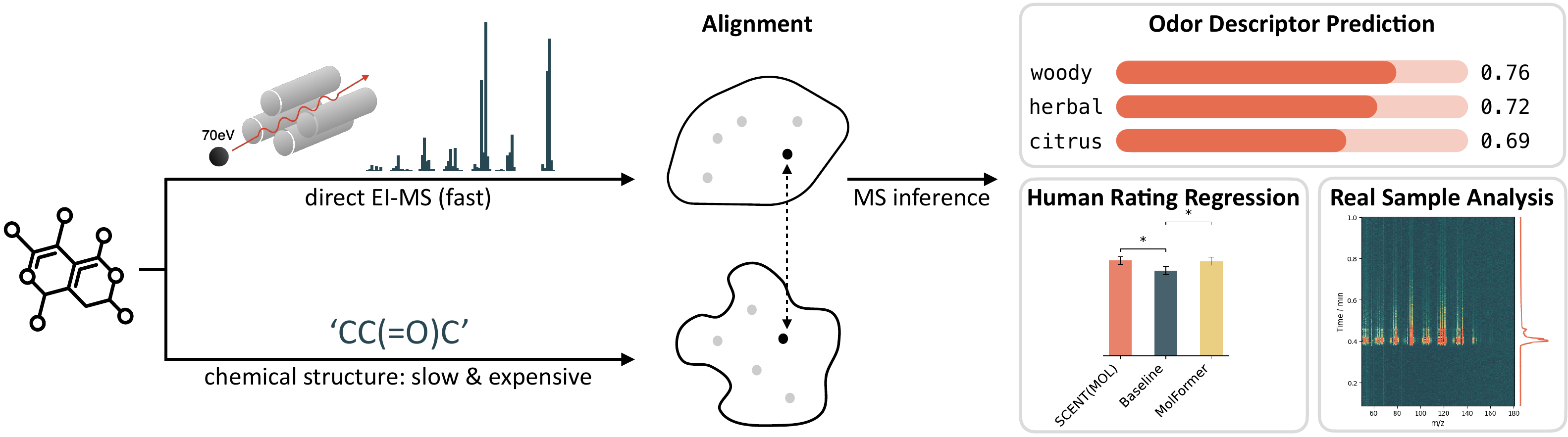}
    \caption{\textbf{Overview of SCENT}. SCENT contrastively aligns EI-MS embeddings with molecular structure embeddings, which are slow and expensive to obtain. At inference, only mass spectra are required. The learned representations are evaluated on three downstream tasks: odor descriptor prediction, human perceptual rating regression, and real-world lab-measured spectra.}
    \label{fig:overview}
    \vspace{-3ex}
\end{figure}

To answer this question, we contribute \textbf{SCENT} (\textbf{S}pectrum-to-\textbf{C}hemical \textbf{E}mbedding alig\textbf{N}men\textbf{T})\footnote{The code and model checkpoints are available at \href{https://github.com/Celestezzz/SCENT}{\texttt{https://github.com/Celestezzz/SCENT}}}, a multi-modal framework for chemically grounded MS representation learning (Fig.~\ref{fig:overview}). SCENT leverages contrastive learning to align MS representations with molecular structure embeddings. Since structure-based olfactory models have shown that molecular representations capture odor-relevant information, this alignment provides a principled way to distill chemical structure information into MS representations, requiring only raw spectra at inference. In particular, SCENT aligns MS embeddings with pretrained molecular structure embeddings using a CLIP-style contrastive objective, enabling the learned MS representations to be used directly for odor prediction.

We evaluate SCENT on odor descriptor prediction and human perceptual rating regression. On a filtered GS-LF benchmark, SCENT substantially improves over MS-only baselines and closes much of the gap to structure-based models, (Section~\ref{odor}). On the DREAM dataset~\citep{dream}, SCENT embeddings better capture human perceptual structure than unaligned MS representations (Section~\ref{dream}). We further analyze data efficiency and provide preliminary validation on real-world, laboratory-measured spectra from 30 single-molecule odorants, showing that the learned representations transfer to practical acquisition conditions with minimal performance degradation (Section~\ref{real}). Together, these results demonstrate that cross-modal alignment with molecular structure enables fast, sensor-derived representations to approximate chemically grounded olfactory reasoning, establishing direct EI-MS as a practical and deployable input modality that bridges rapid chemical sensing and structure-informed odor perception.

\section{Related work}
\textbf{Computational olfaction.} The quantitative structure-odor relationship (QSOR) paradigm is one of the most widely studied frameworks for computational olfaction. \cite{dream} characterizes odorants through 4,884 cheminformatics features (such as molecular weight, functional groups, and chemical bonds) for odor prediction tasks. While these descriptors provide rich molecular information, they rely on manual feature engineering. To address this, recent work has instead adopted SMILES ~\citep{Weininger1988SMILES}, a string-based molecular representation, as input to end-to-end models such as graph neural networks~\citep{sanchez2019machine, Lee2023}. \cite{taleb2024can} further demonstrated that SMILES-based representations can capture odor-related structure under self-supervised learning. All of these approaches, however, assume molecular identity at inference. On the sensor side, \cite{feng2025smellnet} improved E-nose's chemical interpretability by aligning gas sensor array signal with GC-MS. For EI-MS (a faster acquisition modality), existing end-to-end approaches predict odor directly from spectra by treating them as flat feature vectors, discarding their latent chemical structure~\citep{nozaki2016odor, ito2020improvement, hasebe2022exploration}. To the best of our knowledge, we contribute the first computational model that leverages molecular structure at training time to ground EI-MS representations in chemical semantics, while requiring only mass spectra at inference.

\begin{figure}[t]
    \centering
    \includegraphics[width=0.95\linewidth]{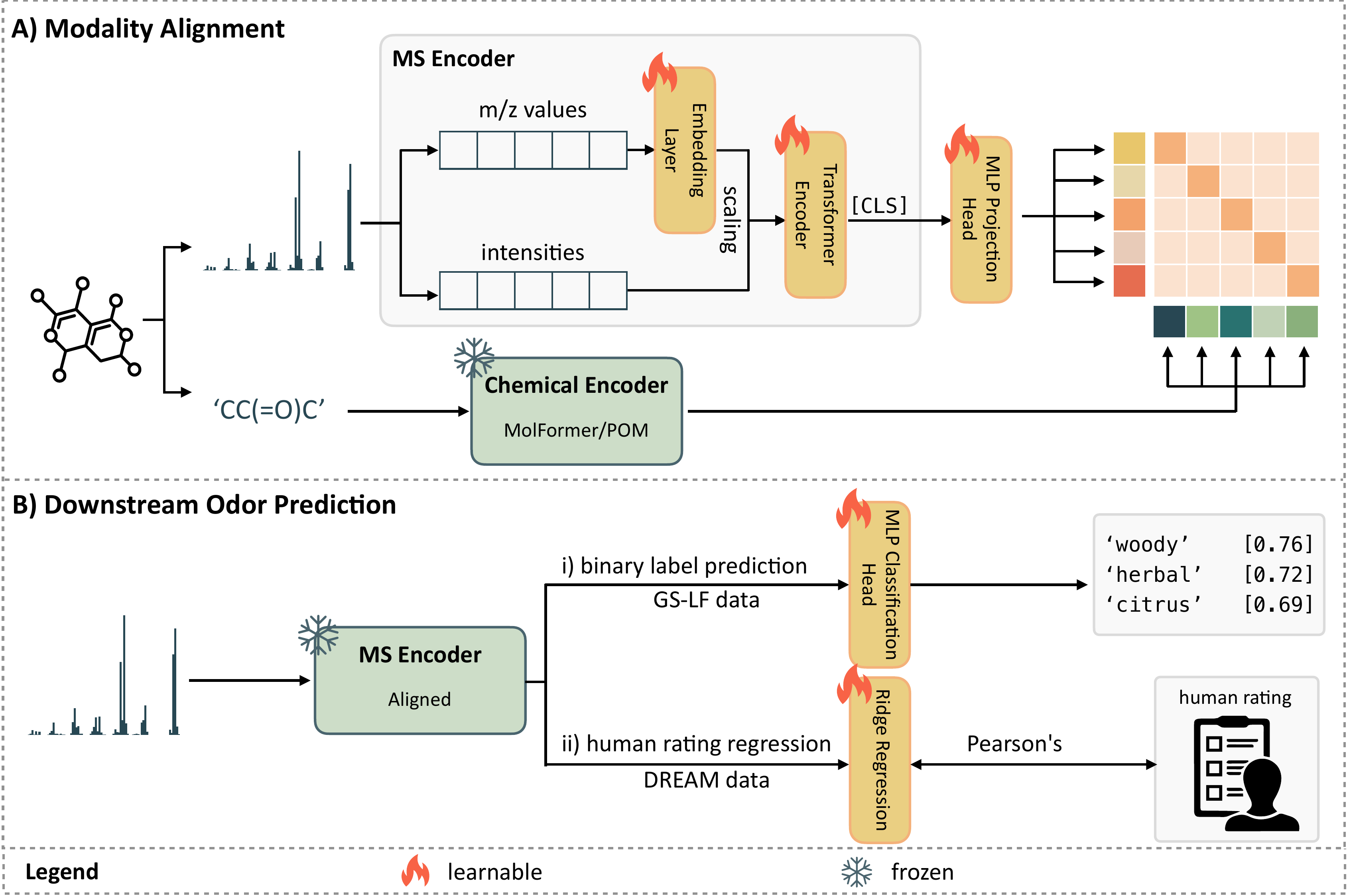}
    \caption{\textbf{The SCENT framework}: \textbf{(A)} A learnable MS encoder and a frozen chemical structure encoder project both modalities into a shared latent space, trained with a multi-modal contrastive loss (Eq.~\ref{eq:pre_loss}); \textbf{(B)} The frozen, aligned MS encoder is evaluated on two downstream tasks: (i) multi-label odor descriptor prediction using a trainable MLP classification head (GS-LF dataset), and (ii) perceptual rating regression using ridge regression (DREAM dataset), evaluated by Pearson correlation. Flame and snowflake icons indicate learnable and frozen components, respectively.}
    \vspace{-3ex}
    \label{fig:network}
\end{figure}

\textbf{Mass spectra representation and contrastive learning.}
The learnability of the relationship between molecular structure and EI-MS has been demonstrated in the forward direction. \citet{wei2019rapid} showed that a convolutional network can predict EI-MS spectra from molecular structure, establishing that structural information is recoverable from spectral patterns.
\citet{10822853} proposed EIMS2Vec, which treats each spectral peak as a token and learns embeddings from peak co-occurrences across a large spectral library, improving the chemical plausibility of EI-MS representations for retrieval tasks. Although such co-occurrence learning can capture recurring fragmentation patterns, it remains driven by spectral statistics alone and does not explicitly ground these patterns in molecular structure. Contrastive learning has emerged as a powerful paradigm for aligning heterogeneous representations across modalities, with notable success in vision and language~\citep{chen2020simple, radford2021learning, zhai2022lit}. It has primarily been applied to align tandem mass spectra, which provide additional fragmentation pathway information, with molecular structure, targeting molecular identification~\citep{goldman2023mist, chen2024cmssp, kalia2025jestr, zhang2026breaking}. Due to the modality difference, these encoders cannot be naturally applied to EI-MS, and their objective is inconsistent with our goal of learning perceptual representations. Instead, in this work we apply contrastive alignment between EI-MS and molecular structure representations, transferring chemical semantics into spectral embeddings for perceptual odor prediction without requiring molecular structure at inference.

\section{Method}
\label{method}
SCENT is a multi-modal contrastive learning framework that aligns EI-MS and structure representations of single-molecule odorants, enabling odor descriptor prediction directly from mass spectra at inference (implementation details and hyperparameters in Appendix~\ref{app:parameter}). As shown in Figure~\ref{fig:network}, SCENT consists of two stages: modality alignment and downstream odor prediction.

\subsection{Modality alignment}
\textbf{Mass spectrum encoder.} A MS represents a molecule as a set of peaks indexed by m/z, an value reflecting the fragment weight, and corresponding intensities (see Appendix~\ref{app:ms} for a details). Inspired by EIMS2Vec~\citep{10822853}, we encode MS data by embedding peak co-occurrences and weighting the embeddings with power-scaled intensities ($p=0.5$). Differently from the original work, we remove the sum pooling over embedding outputs and replace it with a two-layer Transformer encoder with four self-attention heads, which we empirically found to improve performance. The final MS embedding is extracted from a projection MLP applied over the CLS token.

\paragraph{Chemical structure encoder.} To extract chemical embeddings, we consider two pretrained encoders, both kept frozen during alignment training, operating on SMILES~\citep{Weininger1988SMILES} 
representations:
\begin{itemize}
    \item \textbf{Open-POM}~\citep{OpenPOM}, the publicly available version 
    of POM~\citep{Lee2023}, a message passing graph neural network, originally trained with supervision to predict odor labels from molecular structure. To prevent data leakage, we retrain it from scratch on a filtered dataset to predict odor labels from molecular structure, producing a 256-dim embedding per odorant. 
    \item \textbf{MolFormer}~\citep{Ross2022}, a foundation model pretrained 
    on over 1.1 billion molecules, producing a 768-dim embedding per odorant, for which we use the publicly available checkpoints.
\end{itemize}

We refer to the two resulting models as \emph{SCENT (Open-POM)} and \emph{SCENT (MolFormer)} throughout the paper.

\textbf{Training data.}
To train the MS encoder and align MS embeddings with structure embeddings, 
we use:
\begin{itemize}
    \item \textbf{NIST EI-MS library (2023)}: Starting with 347,100 compounds, 
    we filter for molecular weights between 50 and 300 Da (molecule weight unit) to target volatile 
    odorants and remove background noise, resulting in 184,558 unique 
    compound-spectrum pairs.
    \item \textbf{SMILES}: Corresponding SMILES strings are retrieved via 
    PubChem and canonicalized using RDKit, with stereochemistry removed, 
    yielding a final paired dataset of 148,212 compounds for alignment.
\end{itemize}
To prevent data leakage, we ensure that no sample from the odor prediction test set appears in the Open-POM retraining or in the alignment training datasets. The details are described in Appendix~\ref{app:workflow}.

\textbf{Training objective.} SCENT aligns MS and structure embeddings through a contrastive objective 
inspired by CLIP~\citep{radford2021learning}. Within a mini-batch of $N$ 
pairs, we compute a similarity matrix using temperature-scaled cosine 
similarity $s_{ij} = (u_{i}^T v_j)$. The training objective minimizes the symmetric 
alignment loss:
\begin{equation}
\label{eq:pre_loss}
\mathcal{L}_{\text{align}} = - \frac{1}{2N} \sum_{i=1}^{N} \left( 
\log \frac{e^{s_{ii}/\tau}}{\sum_{j=1}^{N} e^{s_{ij}/\tau}} + 
\log \frac{e^{s_{ii}/\tau}}{\sum_{j=1}^{N} e^{s_{ji}/\tau}} \right),
\end{equation}
where $\tau$ is a temperature parameter selected via grid search on validation AUC, which is also used to determine the early stopping point.

\subsection{Downstream odor prediction}
\label{sec:donwstream}

We evaluate SCENT on two downstream tasks that probe complementary aspects of olfactory perception: multi-label odor descriptor prediction, which tests the ability of the learned representations to capture discrete perceptual categories, and human perceptual rating regression, which assesses how well the aligned MS representations approximate continuous human odor judgments.
\subsubsection{Multi-label odor descriptor prediction}
\textbf{Dataset and protocol.}
Following~\citep{Lee2023}, we evaluate the learned representations on odor classification using the GS-LF dataset~\citep{gs, lf}, which contains 138 odor descriptor labels for single-molecule odorants. We use 2,588 single-molecule samples with valid MS spectra in the filtered NIST library for downstream evaluation. To obtain a robust evaluation protocol, we first construct a fixed test set comprising 10\% of the samples using iterative stratified splitting, which preserves the label co-occurrence distribution. The remaining 90\% forms the training-validation pool, on which we perform 5-fold iterative stratified cross-validation. To prevent data leakage, any alignment training sample whose MS spectrum is identical to a downstream test sample is removed from the alignment pool. On top of the learned embeddings, we train a simple 2-layer MLP classifier using a weighted binary cross-entropy loss. The label distribution for each fold is shown in Appendix~\ref{app:label}.

\textbf{Metrics.} We report the following evaluation metrics:
\begin{itemize}
    \item \textbf{ROC-AUC Score}: We evaluate model performance using both micro-averaged AUC and weighted-AUC. Micro-AUC aggregates predictions across all classes and measures overall discriminative performance, and is used to select the early stopping point. Weighted-AUC computes per-class AUC and averages them with weights proportional to class frequency, providing a more balanced assessment under class imbalance and long-tailed label distributions. The labels with no samples in test set will not be counted. Validation AUC is used for selecting early stopping point.
    \item \textbf{Precision@k}: We adopt an adjusted Precision@k metric to evaluate the proportion of true positive labels among the top-$k$ predictions produced by the model. Since some samples may contain fewer than $k$ positive labels, we define:
\begin{equation}
    \mathrm{Adj.\ Precision@k} = \frac{\mathrm{hit@k}}{\min(k, n_i^+)},
\end{equation}
    where $n_i^+$ denotes the number of ground-truth positive labels for the $i$-th sample, and $hit@k$ is the number of correctly predicted labels in the top-$k$ predictions. In this work we use $k=5$.
\end{itemize}

\subsubsection{Human perceptual rating regression}
\textbf{Dataset and protocol.}
We use the DREAM olfaction dataset~\citep{dream}, which contains perceptual ratings from 49 non-expert participants on a scale of 0--99 across 21 descriptors for 476 molecules. The dataset includes measurements at three dilution levels, although not every molecule was evaluated at all concentrations. We filter the dataset to retain only molecules with valid MS spectra in the library, leaving 385 molecules. Since MS representations do not encode concentration, we select the highest available concentration for each molecule, average ratings across participants, and normalize scores to [0,1]. We evaluate using 100 repeated train-test splits with an 80/20 ratio, fitting a ridge regression model on frozen MS embeddings to predict 21 continuous perceptual attributes.

\begin{table}[t]
\centering
\caption{\label{tab:results} \textbf{Model performance on the filtered GS-LF dataset}. Results are reported as mean $\pm$ std over 5 folds. Structure-based models (Open-POM, MolFormer) serve as upper-bound references, EIMS2Vec as the MS-only baseline, and SCENT as our proposed method. {\color{teal}$\uparrow$} denotes improvement over EIMS2Vec. Bold indicates the best result per metric.}
\vspace{1ex}
\resizebox{\columnwidth}{!}{%
\begin{tabular}{llccc}
\toprule
\textbf{Model} & \textbf{Input} &
\textbf{Micro-AUC}$\uparrow$ &
\textbf{Weighted-AUC}$\uparrow$ &
\textbf{Adj.P@k}$\uparrow$ \\
\midrule
Open-POM  & SMILES & $89.94\pm0.05$ & $\mathbf{86.70\pm0.15}$ & $\mathbf{52.22\pm0.28}$ \\
MolFormer & SMILES & $89.29\pm0.20$ & $84.14\pm0.44$          & $50.23\pm0.16$ \\
\midrule
EIMS2Vec  & MS     & $87.98\pm0.15$ & $82.29\pm0.61$          & $46.49\pm0.53$ \\
\midrule
\textbf{SCENT (Open-POM)}
& MS
& $89.80\pm0.09$ \vsbase{1.82} 
& $85.45\pm0.30$ \vsbase{3.16} 
& $50.60\pm0.53$ \vsbase{4.11} \\
\textbf{SCENT (MolFormer)}
& MS
& $\mathbf{89.99\pm0.18}$ \vsbase{2.01} 
& $85.71\pm0.26$ \vsbase{3.42}
& $50.62\pm0.75$ \vsbase{4.13} \\
\bottomrule
\end{tabular}
}
\vspace{-3ex}
\end{table}

\textbf{Metric.}
Following prior work~\citep{taleb2024can}, we use the Pearson correlation coefficient ($r$) between predicted and human-rated perceptual scores to evaluate how well the aligned MS representations approximate human perceptual judgments.

\section{Results}
We report quantitative results for odor descriptor prediction, including ablation and data efficiency analyses (Section~\ref{odor}), human perceptual rating regression (Section~\ref{dream}), and validation on real-world lab-measured spectra (Section~\ref{real}). We provide additional results on in Appendix~\ref{app:add_result_1},~\ref{app:add_result_2} and~\ref{app:add_result_3}.

\subsection{Odor descriptor prediction}
\label{odor}
\textbf{Main results.}
Model performance is shown in Table \ref{tab:results}. Both SCENT variants significantly improve over the unaligned MS embedding, demonstrating the effectiveness of cross-modal alignment. Notably, SCENT achieves performance comparable to the structure-based models despite relying solely on MS data at inference time, suggesting that alignment enables the MS encoder to learn representations that capture chemically relevant information and serve as a surrogate for explicit chemical structure in downstream tasks. We also visualize the learned representation space using PCA and show in Figure~\ref{fig:PCA_full}. The relationship between intra-label spectral/structural similarity and per-label AUC gain in Appendix~\ref{fig:similarity}, confirming that SCENT's advantage is consistent across descriptors regardless of label frequency or data similarity.

To understand the source of the residual weighted-AUC, we group descriptors by label frequency and compare mean per-label AUC within each bin (Figure~\ref{fig:mean_label_auc_mol} and Figure~\ref{fig:mean_label_auc_pom} in Appendix). SCENT consistently outperforms the MS-only baseline across all frequency groups, with the largest gains in low- and medium-frequency bins. Across bins, SCENT remains broadly comparable to the corresponding structure-based models, suggesting that the residual weighted-AUC gap is not driven by one specific frequency regime failure, but reflects modest differences accumulated across label groups.
\begin{figure}[t] %
    \centering
    \includegraphics[width=\linewidth]{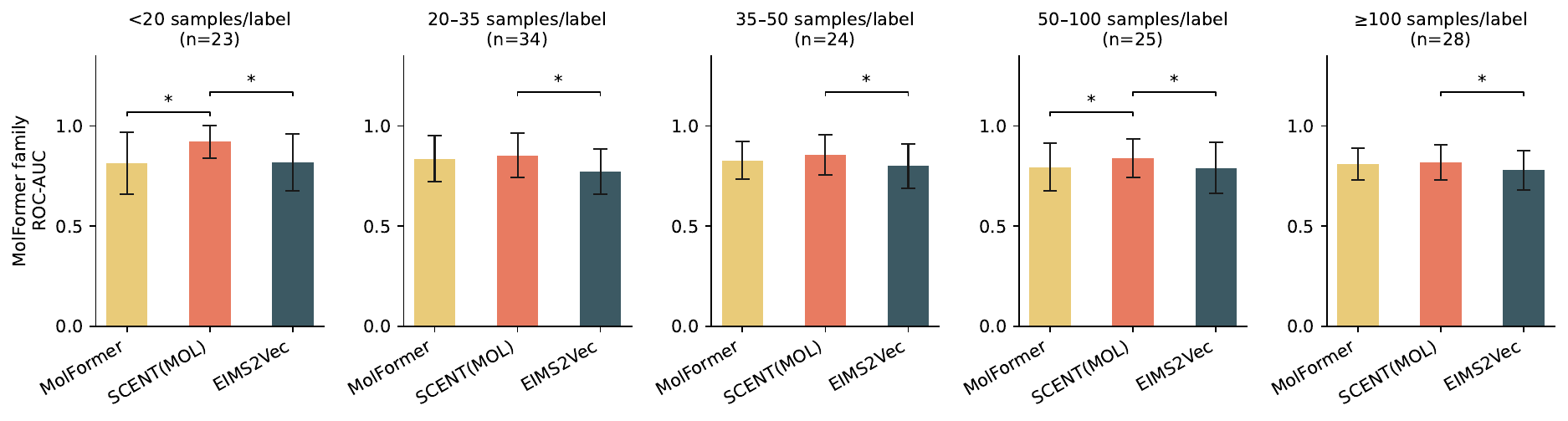}
    \caption{\textbf{Mean per-label ROC-AUC grouped by label frequency (MolFormer-based models).} Each bar reports the mean AUC across labels within a given frequency bin (error bars indicate standard error of the mean). SCENT (MolFormer) consistently outperforms EIMS2Vec across all frequency bins, with statistically significant gains even in low-frequency categories. Statistical significance is assessed via the Wilcoxon signed-rank test (* $p < .05$). More results in Appendix~\ref{app:results:label_auc}.}
    \label{fig:mean_label_auc_mol}
    \vspace{-3ex}
\end{figure}

\begin{table}[b]
\centering
\caption{\label{tab:ablation} \textbf{Ablation study of SCENT} (mean $\pm$ std over 5 folds). We ablate two components: the Transformer head and the training objective. \textit{w/o CLIP 1} replaces the contrastive objective with masked reconstruction (MR); \textit{w/o CLIP 2} replaces it with end-to-end supervised training (S); \textit{w/o Transformer} retains contrastive alignment (C) but removes the Transformer head. EIMS2Vec uses in-spectrum peak co-occurrence (PC) and serves as the baseline.}
\vspace{1ex}
\resizebox{\columnwidth}{!}{%
\begin{tabular}{lccccc}
\toprule
\textbf{Model Variant}  & 
\textbf{Transformer} & \textbf{Objective} & \textbf{Micro-AUC}$\uparrow$ & \textbf{Weighted-AUC}$\uparrow$ & \textbf{Adj.Precision@k}$\uparrow$\\
\midrule
w/o CLIP 1                   & \checkmark & MR & $76.34\pm0.03$ & $55.39\pm0.45$ & $29.95\pm0.20$\\
w/o CLIP 2                   & \checkmark & S  & $77.90\pm0.22$ & $64.60\pm0.67$ & $29.99\pm0.10$\\
w/o Transformer (Open-POM)  & \texttimes & C  & $88.39\pm0.12$ & $83.04\pm0.60$ & $47.40\pm0.59$\\
w/o Transformer (MolFormer) & \texttimes & C  & $88.40\pm0.08$ & $83.15\pm0.34$ & $47.42\pm0.59$\\
EIMS2Vec                    & \texttimes & PC & $87.98\pm0.15$ & $82.29\pm0.61$ & $46.49\pm0.53$ \\
\bottomrule
\end{tabular}
}
\vspace{-3ex}
\end{table}

\textbf{Ablation study.}
\label{app:ablation}
To understand the contribution of each component of SCENT, we conduct an ablation study varying the presence of the Transformer head and the contrastive alignment objective (Table~\ref{tab:ablation}). When both components are removed, the model reduces to EIMS2Vec, which serves as our baseline. All variants are trained following the same experimental settings, using MS data as the sole input. When the contrastive alignment objective is replaced with a BERT-like masked reconstruction objective~\citep{devlin2019bert} with a 15\% masking ratio (w/o CLIP 1), performance drops substantially, indicating that SCENT's gains cannot be attributed to spectrum-only pretraining or encoder capacity alone. Similarly, replacing the contrastive objective with direct end-to-end supervised training on odor labels (w/o CLIP 2) yields comparably poor performance, confirming that neither the encoder capacity nor the availability of label supervision alone is sufficient. It is specifically the alignment with pretrained chemical structure embeddings that transfers chemically grounded information into the MS representation space. Removing the Transformer head while retaining alignment yields a smaller but consistent degradation, suggesting that the Transformer improves the modeling of peak co-occurrence patterns in MS. Together, these results confirm that the dominant source of improvement is the alignment of MS representations with chemically informative structure embeddings, with the Transformer providing a complementary but secondary contribution.

\begin{figure}[t] %
    \centering
    \includegraphics[width=1\linewidth]{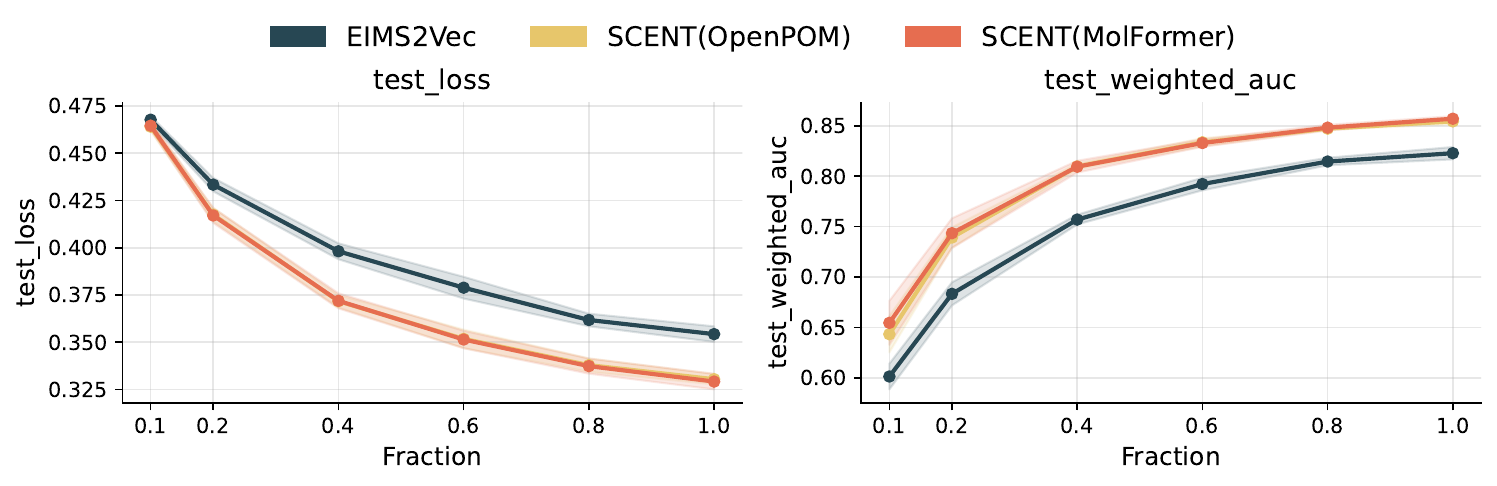}
    \caption{\textbf{Test performance vs.\ training set fraction.} Test loss (left) and weighted-AUC (right) are reported for SCENT (OpenPOM), SCENT (MolFormer), and EIMS2Vec across varying fractions of the training set (0.1 to 1.0), as mean $\pm$ std over 5 folds. SCENT consistently outperforms EIMS2Vec across all fractions. More results in Appendix~\ref{app:data_scaling}.}
    \label{fig:test_vs_fraction_1} 
    \vspace{-1ex}
\end{figure}

\textbf{Dataset scaling}.
To evaluate the data efficiency of SCENT, we train all models on varying fractions of the training set (from 10\% to 100\%) and report test performance across 5 folds (Figure~\ref{fig:test_vs_fraction_1} and Figure~\ref{fig:test_vs_fraction_2}). Both SCENT variants consistently outperform EIMS2Vec across all data regimes and metrics. While all models improve as the training fraction increases, SCENT maintains a clear and consistent advantage, suggesting that multi-modal alignment pretraining instills a chemical prior that the MS encoder cannot recover from spectral signal alone, and that the resulting representations generalize well even in the low-data regime.

\begin{figure}[t]
    \centering
    \includegraphics[width=1.0\linewidth]{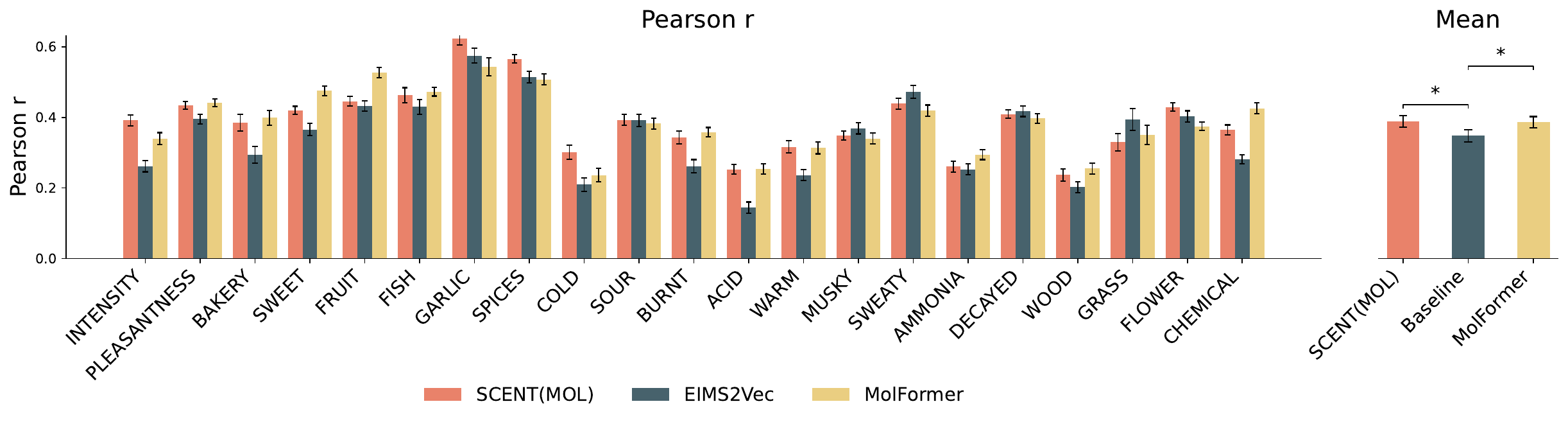}
    \caption{\textbf{Pearson correlation between predicted and human-rated perceptual scores on the DREAM dataset (MolFormer family).} Ridge regression is fitted on frozen embeddings to predict 21 continuous perceptual descriptors (mean $\pm$ std across 100 folds). SCENT (MolFormer) significantly outperforms the unaligned baseline and achieves performance comparable to MolFormer despite requiring no molecular structure at inference. Statistical significance is assessed via the Wilcoxon signed-rank test~\citep{wilcoxon1992individual} (* $p<.05$). More results in Appendix~\ref{app:human_reg:stats_tests}.}
    \label{fig:keller_r_mol}
    \vspace{-2ex}
\end{figure}

\subsection{Human perceptual rating regression}
\label{dream}
To better understand the trend similarity between model prediction and human rating, Figure~\ref{fig:keller_r_mol} and Figure~\ref{fig:keller_r_pom} shows the Pearson correlation coefficient for each descriptor between regression predictions and human ratings.  Both SCENT variants outperform the unaligned EIMS2Vec baseline, indicating that spectrum-structure alignment improves the ability of MS-derived representations to capture human perceptual ratings, transferring perceptual grounding that is otherwise a key property of structure-based representation.

SCENT (MolFormer) achieves performance comparable to MolFormer, despite using only MS spectra at inference time. Since MolFormer embeddings are not trained with explicit olfactory supervision, this result indicates that general chemical structure representations already encode information relevant to human perception, and that SCENT can effectively transfer this information into the MS embedding space. In contrast, SCENT (Open-POM) remains below Open-POM, which is expected: Open-POM embeddings are directly optimized with odor supervision and therefore contain task-specific information that cannot be fully recovered from MS signals alone. It should be noticed that the statistical test is applied for each label pair, not simply on the averaged and standard deviation values.

\subsection{Case study: real-world chemical spectra}
\label{real}
Deploying the model on real-world measurements introduces a domain gap relative to library references: relative peak intensities vary with instrument conditions, residual background noise introduces spurious low-intensity peaks, and the absence of chromatographic separation may allow impurity to corrupt the spectral profile. To evaluate robustness to this domain gap, we conduct experiments on lab-measured MS data collected from 30 single-molecule odorants, adapting the model through (1) a dedicated signal pre-processing pipeline, (2) re-trained under a unified m/z range, and (3) evaluation on both library data and real-world data.

\textbf{Model adaptation.}
To mitigate the influence of air-related ions in the low m/z region (small molecules) and low-volatility fragments in the high m/z region (heavy molecules), real measurements were restricted to the m/z range of [50, 180]. We fine-tune the model using the same m/z range consistently across NIST library data, ensuring alignment between training and inference conditions. The fine-tuned model is then evaluated on the same downstream tasks and datasets described in Section~\ref{sec:donwstream}, with results reported in Appendix (Table ~\ref{tab:50180_results}).

\begin{figure}[t] %
    \centering
    \includegraphics[width=1\linewidth]{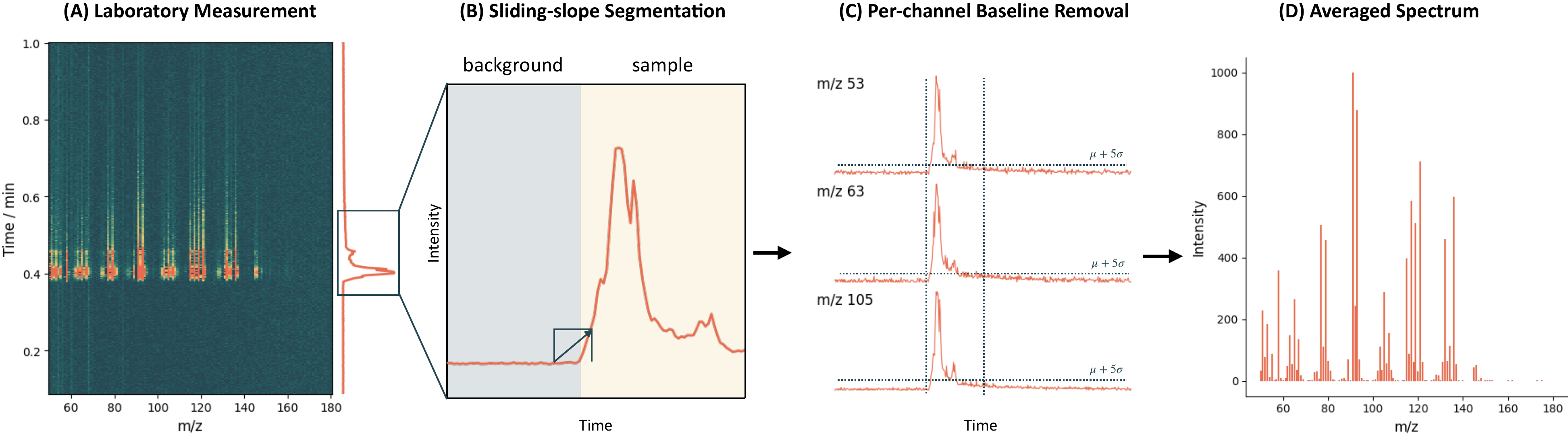}
    \caption{\textbf{Signal pre-processing pipeline for lab-collected spectra.} \textbf{(A)} Raw ion intensity is collected as a 2D matrix over m/z channels and time. \textbf{(B)} A sliding-slope algorithm segments the total ion current into background and sample regions. \textbf{(C)} Per-channel baseline correction uses a noise threshold of $\mu_\text{bg}$ + $5\sigma_\text{bg}$ estimated from the background region. \textbf{(D)} Corrected intensities within the sample window are averaged over time to produce a single spectrum as model input.}
    \label{fig:sigal_processing}
    \vspace{-3ex}
\end{figure}

\textbf{Data collection and pre-processing.} The raw MS signals are collected in laboratory air, comprising a background period (about 0.3 min) followed by sample exposure (about 0.7 min), with an overall measurement duration of around 60 seconds. For each time point, we compute the total ion intensity by summing signals across all m/z channels, forming a temporal intensity trace that reflects the sample elution profile. We apply a sliding-slope algorithm with a 20-time point window to this temporal intensity trace to identify key boundaries of the sample region, where the steepest increase indicates the onset of sample elution and the subsequent increase of slope marks the tailing start and signal termination points. This procedure separates the sample window from background regions. For each m/z channel, we estimate noise statistics by computing the mean $\mu_{\mathrm{bg}}$ and standard deviation $\sigma_{\mathrm{bg}}$ over time. Baseline correction is then performed by subtracting a threshold defined as $\mu_{\mathrm{bg}} + 5\sigma_{\mathrm{bg}}$, which suppresses background noise while preserving statistically significant ion signals. Finally, the corrected intensities within the sample window are averaged over time to form the model input. A summary of this procedure is depicted in Figure~\ref{fig:sigal_processing}.

\textbf{External test set.}
We selected 30 samples (28 from the filtered GS-LF test set, the remaining two with labels sourced from~\cite{gs}) and conducted laboratory air measurements (we present the full list of samples in Appendix~\ref{app:case_study:dataset}). To quantify uncertainty arising from the small real-world validation set, we performed non-parametric bootstrap resampling over these 30 molecules. For each bootstrap replicate, we sampled 30 molecules with replacement, recomputed metrics for each of the five classification heads, and averaged the metric across heads. We repeated this procedure 10,000 times and reported the 2.5$th$ and 97.5$th$ percentiles as the 95\% bootstrap confidence interval. The results are shown in Table~\ref{tab:real_results}. The bootstrap intervals indicate that SCENT's real-world advantage is most reliable for Micro-AUC and Adj.P@5, suggesting better top-ranked odor descriptor retrieval from lab-measured spectra. Weighted-AUC shows smaller differences with overlapping confidence intervals, indicating that this metric remains more sensitive to the limited 30-sample external test set. SCENT evaluated on real-world spectra still outperforms EIMS2Vec evaluated on clean library spectra across all metrics, confirming that the advantage of multi-modal pre-training is preserved under realistic measurement conditions.

\begin{table}[t]
\centering
\caption{\label{tab:real_results}
\textbf{Model performance on real-world lab-measured spectra} (30 single-molecule odorants). Results are reported as mean $\pm$ std across five classification heads, with 95\% bootstrap confidence intervals obtained by resampling the 30 test molecules. Bold indicates the best result per metric.
}
\vspace{1ex}
\resizebox{\columnwidth}{!}{%
\begin{tabular}{llccc}
\toprule
\textbf{Metric} & \textbf{Statistic} &
\textbf{EIMS2Vec} &
\textbf{SCENT (Open-POM)} &
\textbf{SCENT (MolFormer)} \\
\midrule
\multirow{2}{*}{Micro-AUC$\uparrow$}
  & mean $\pm$ std & $84.80 \pm 0.52$ & $\mathbf{88.67 \pm 0.49}$ & $87.75 \pm 0.67$ \\
  & 95\% CI        & $[82.63, 86.96]$ & $\mathbf{[86.94, 90.36]}$ & $[85.49, 89.79]$ \\
\midrule
\multirow{2}{*}{Weighted-AUC$\uparrow$}
  & mean $\pm$ std & $78.46 \pm 0.80$ & $\mathbf{79.42 \pm 0.49}$ & $78.92 \pm 0.82$ \\
  & 95\% CI        & $[73.98, 82.46]$ & $\mathbf{[75.24, 83.46]}$ & $[75.20, 82.28]$ \\
\midrule
\multirow{2}{*}{Adj.P@5$\uparrow$}
  & mean $\pm$ std & $40.67 \pm 1.80$ & $\mathbf{51.18 \pm 1.01}$ & $49.93 \pm 1.39$ \\
  & 95\% CI        & $[33.00, 48.53]$ & $\mathbf{[44.58, 58.27]}$ & $[42.17, 57.83]$ \\
\bottomrule
\end{tabular}
}
\vspace{-3ex}
\end{table}

\section{Discussion}
We introduced SCENT, a cross-modal contrastive learning framework that injects chemical structure information into MS-based representations at training time while requiring only mass spectra at inference. Our results show that aligning direct EI-MS signals with pretrained chemical structure embeddings substantially improves odor descriptor prediction over an MS-only baseline, closing most of the gap to structure-based models without requiring any structural information at test time. The emergence of perceptually coherent clusters in the aligned representation space, and the consistent advantage of SCENT under reduced training set sizes, together suggest that cross-modal alignment transfers chemically meaningful structure into the spectral embedding space as a direct consequence of the alignment objective, independent of any downstream supervision. Beyond controlled library settings, we validated SCENT on lab-measured spectra collected from real odorant samples, demonstrating that the learned representations transfer to real-world acquisition conditions without any domain-specific fine-tuning. Despite the domain gap introduced by residual background noise and the absence of chromatographic separation, SCENT maintained competitive performance, suggesting that the chemically grounded representations learned through alignment are robust to the signal distortions inherent in direct sampling scenarios.

\textbf{Limitations.} First, both alignment training and downstream evaluation rely on library-acquired spectra, which differ from real-world measurements in peak intensity distributions and background contamination. While our case study demonstrates encouraging transfer to lab-measured signals, the domain gap between library and real-world spectra remains a key challenge for deployment, and the small size of our real-world test set (30 samples) limits the strength of conclusions that can be drawn. Second, the current framework assumes single-molecule inputs; extending alignment to mixed-compound spectra, which are common in direct sampling settings, is a non-trivial open problem. Third, our framework does not explicitly model odorant concentration. While mass spectrometric measurements may contain abundance-related information, the representation used here mainly reflects relative fragmentation structure. Consequently, the model cannot capture concentration-dependent and potentially non-linear changes in human odor perception.

\textbf{Future work.} Future work will focus on three directions: domain adaptation to bridge the library-to-real-world gap, augmentation strategies for low-support odor labels, and extending the alignment framework to mixed-signal settings where chromatographic separation is unavailable. More broadly, we believe that cross-modal alignment between analytical sensing modalities and molecular structure representations offers a promising direction for grounding rapid chemical measurements in chemically and perceptually meaningful spaces.

\section{Acknowledgments}
This work was supported by the Knut and Alice Wallenberg Foundation, Swedish Research Council, ERC AdV BIRD 88480, and ERC Syn D2Smell 101118977. The computational experiments were enabled by the Berzelius resource provided by the Knut and Alice Wallenberg Foundation at the National Supercomputer Centre.

We would thank to Jiawei Li and David Vävinggren for feedback on early version of this manuscript.

\bibliographystyle{abbrvnat}
\bibliography{cite}


\newpage
\appendix
\setcounter{table}{0}
\setcounter{figure}{0}
\counterwithin*{table}{section}
\counterwithin*{figure}{section}
\renewcommand{\thetable}{\thesection\arabic{table}}  
\renewcommand{\thefigure}{\thesection\arabic{figure}}

\section{SCENT workflow}
\label{app:workflow}
This section summarizes the experimental workflow. Detailed model architecture, objectives, and hyperparameters are provided in Section~\ref{method} and Appendix~\ref{app:parameter}.

\textbf{Data division.}
We filter GS-LF dataset with molecules weight in 50-300 Da, remains 2,588 molecule. We first split the 2,588 molecule-spectrum pairs with valid MS spectra into a fixed 10\% test set and a 90\% training-validation pool using iterative stratified splitting of order 2~\citep{OpenPOM}. Test set molecules are excluded from all upstream retraining and alignment data to avoid leakage.

\textbf{Chemical encoder preparation.}
We use two chemical structure encoders as alignment targets: a leakage-controlled Open-POM model retrained after removing downstream test molecules from full GS-LF dataset, and the public MolFormer checkpoint used without fine-tuning. These encoders provide fixed molecular embeddings from SMILES strings.

\textbf{Alignment data construction.}
SMILES strings are retrieved from PubChem using CAS identifiers and canonicalized with RDKit after removing stereochemistry
(\texttt{canonical=True, isomericSmiles=False}). After discarding compounds without valid SMILES, 148,212 compound--spectrum pairs remain for alignment; duplicate canonical SMILES are retained as independent spectral entries.

\textbf{EIMS2Vec spectrum embedding.}
For a spectrum, each observed peak $j$ is represented by the corresponding center embedding $\mathbf{e}_{\text{cen},j}$, while peak intensity controls its contribution through power scaling. The resulting EIMS2Vec spectrum embedding is
\begin{equation}
\mathbf{u} = \sum_j \tilde{x}_j^{0.5} \cdot \mathbf{e}_{\text{cen},j},
\end{equation}
where $\tilde{x}_j$ is the normalized intensity of peak $j$ and $p=0.5$ compresses the dynamic range.

\textbf{Spectrum-structure alignment.}
SCENT initializes spectral peak representations from $E_{\text{cen}}$ without the sum pooling. The embedding will be add a learnable \texttt{[CLS]} token at the head, and send to a Transformer head with $2$ layers with $4$ self-attention head. The final \texttt{[CLS]} will be linearly projected (learnable) into same dimension with corresponding chemical embedding from the frozen Open-POM or MolFormer model. 

\section{Hardware}
All computational experiments were conducted on a computing node equipped with an AMD EPYC 7502 32-Core Processor and an NVIDIA Quadro RTX 6000 GPU (24GB VRAM). The models were implemented using the PyTorch framework and optimized with the Adam optimizer. The alignment training on full NIST2023 dataset takes approximately 1 hour per run. The odor prediction task takes 0.5 hour for 5 folds. The linear regression training doesn't need GPU source.
\section{Hyperparameters}
We use validation loss during contrastive alignment to determine the early stopping point and to select the optimal temperature parameter $\tau$. The full hyperparameter configuration for each stage of SCENT is reported in Table~\ref{tab:hyper}.
\label{app:parameter}
\begin{figure}[h] %
    \centering
    \includegraphics[width=0.8\linewidth]{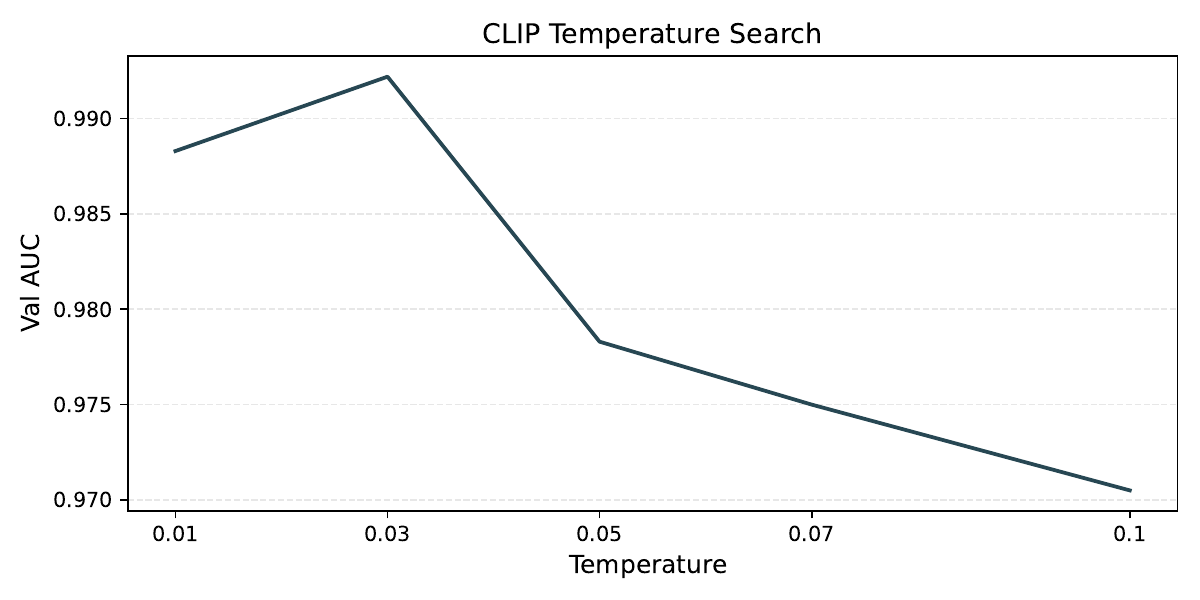}
    \caption{\textbf{Contrastive learning performance  with different temperature $\tau$.} When $\tau=0.03$, the model performances the best on alignment.}
    \label{fig:tem} 
\end{figure}
\begin{table}[h]
    \centering
    \caption{Hyperparameter settings for training and downstream task.}
    \vspace{1ex}
    \label{tab:hyper}
    \renewcommand{\arraystretch}{1.2}
    \begin{tabular}{llcc}
        \toprule
        \textbf{Module} & \textbf{Hyperparameter} & \textbf{Symbol} & \textbf{Value} \\
        \midrule
        \multirow{5}{*}{MS Encoder} 
        & Transformer Layers & $L$ & 2 \\
        & Attention Heads & $H$ & 4 \\
        & Embedding Dimension & $d$ & 500 \\ 
        & Dropout Rate & $p$ & 0.1 \\
        & Activation & - & ReLU\\
        \midrule
        \multirow{6}{*}{Alignment} 
        & Batch Size & $B$ & 512 \\
        & Encoder Learning Rate & $\eta_{enc}$ & 1e-4 \\
        & Projection Learning Rate & $\eta_{proj}$ & 1e-3 \\
        & Optimizer & - & Adam \\
        & Temperature & $\tau$ & 0.03 \\
        & Max Epochs & - & 50 \\
        \midrule
        \multirow{4}{*}{Downstream} 
        & Batch Size & $B$ & 64 \\
        & MLP Learning Rate & $\eta_{cla}$ & 1e-4 \\
        & w/o CLIP 2 Encoder Learning Rate & $\eta_{enc_a}$ & 1e-4 \\
        & MLP Hidden Dim Ratio & $r_{mlp}$ & 0.5 \\
        & Max Epochs & - & 50 \\
        \bottomrule
    \end{tabular}
\end{table}
\section{Introduction of signal modality}
\label{app:ms}
In this work, we focus on direct EI-MS. Our choice is motivated by its balance between interpretability with real-time accessibility for computational olfaction. 

A mass spectrum represents a molecule as a discrete set of peaks, each defined by two quantities: its mass-to-charge ratio (m/z), an integer index reflecting the mass of a molecular fragment produced during ionization, and its intensity, which indicates the relative abundance of that fragment. Together, the pattern of peaks encodes a molecule's fragmentation fingerprint, which is a structured chemical signature that is reproducible across instruments and directly comparable to reference libraries such as NIST. In this section, we compare direct EI-MS with other sensing modalities and highlight the key rationale behind our selection.

\begin{itemize}
    \item \textbf{Hard vs soft ionization}: Compared to online mass spectrometric techniques such as proton-transfer-reaction mass spectrometry (PTR-MS), which rely on softer ionization and often allow only compound-class level assignments, EI-based single-quadrupole MS enables direct comparison to established spectral libraries like NIST library. This allows us to have a certain degree toward molecular-level interpretability, with substantially reduced instrumental complexity and cost. 
    \item \textbf{Time vs separation}: Unlike GC-EI-MS, which requires time-consuming separation with experts' experience (in the order of hours), direct EI-MS enables rapid acquisition (in the order of seconds).
    \item \textbf{Electronic sensor vs analytical equipment:} In the realm of rapid olfactory sensing, gas sensor arrays are widely adopted. However, they offer only non-specific digital signal fingerprints and lack explicit chemical meaning~\citep{feng2025smellnet}. In contrast, analytical instrumentation offers molecular-level information, establishing a rigorous chemical basis for olfactory computation.
\end{itemize}

\section{Additional results: odor descriptor prediction }
\label{app:add_result_1}
\subsection{Label grouped AUC}
\label{app:results:label_auc}
\begin{figure}[t] %
    \centering
    \includegraphics[width=1\linewidth]{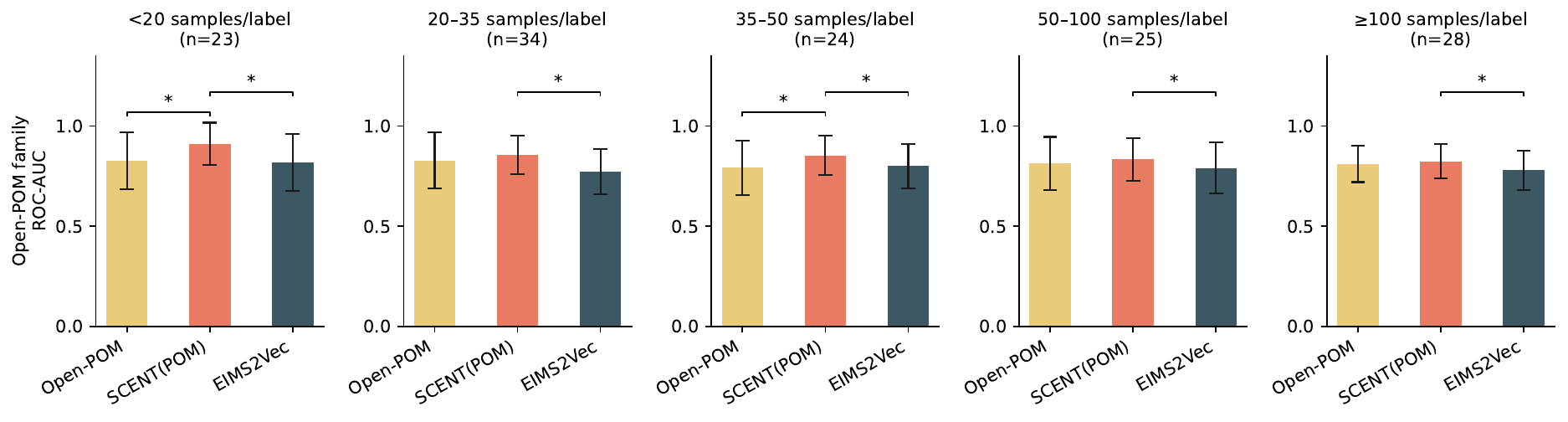}
    \caption{\textbf{Mean per-label ROC-AUC grouped by label frequency for Open-POM based models.} Each bar reports the mean AUC across labels within a given frequency bin, with a standard error of the mean error bar. Statistical significance is assessed via the Wilcoxon signed-rank test (* $p < .05$).}
    \label{fig:mean_label_auc_pom} 
\end{figure}

We further analyze model performance stratified by label frequency in test set to understand where SCENT's gains are concentrated. Figure \ref{fig:mean_label_auc_mol} and Figure~\ref{fig:mean_label_auc_pom}  reports mean per-label ROC-AUC within five frequency bins, evaluated using the Wilcoxon signed-rank test for pairwise significance \citep{wilcoxon1992individual}. The results show that SCENT consistently outperforms EIMS2Vec across all bins, with the largest absolute gains in low- and medium-frequency categories, suggesting that chemical structure alignment is particularly beneficial when training signal per label is scarce. There are 134 labels, while 4 labels has no samples in test set (see Figure~\ref{fig:label_distribution_full}).

\subsection{Model performance and intra-group similarity}
\begin{figure}[h]
    \centering
    \begin{subfigure}[b]{0.48\linewidth}
        \centering
        \includegraphics[width=\linewidth]{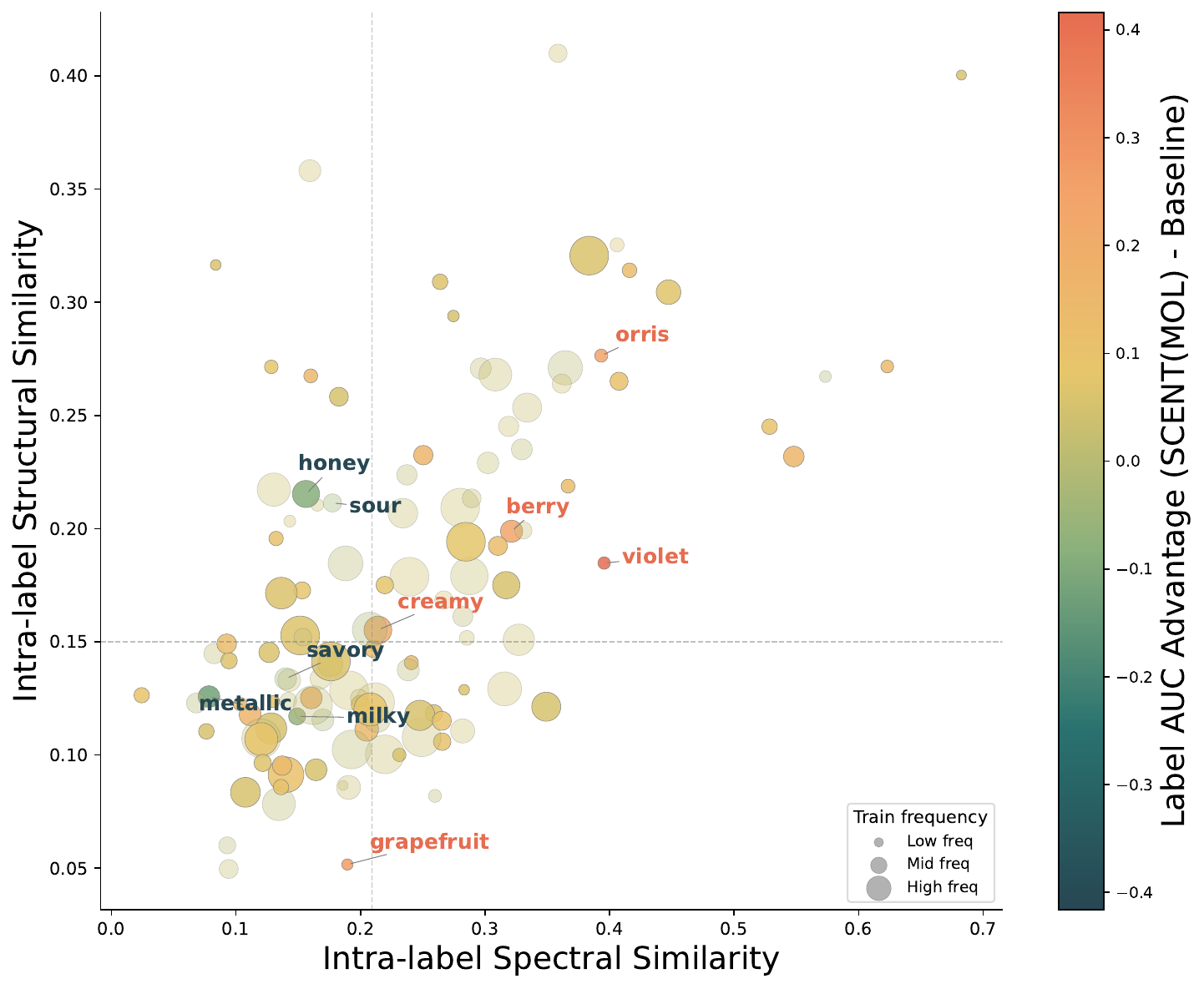}
        \caption{SCENT (MolFormer) vs. EIMS2Vec}
        \label{fig:similarity_mol}
    \end{subfigure}
    \hfill
    \begin{subfigure}[b]{0.48\linewidth}
        \centering
        \includegraphics[width=\linewidth]{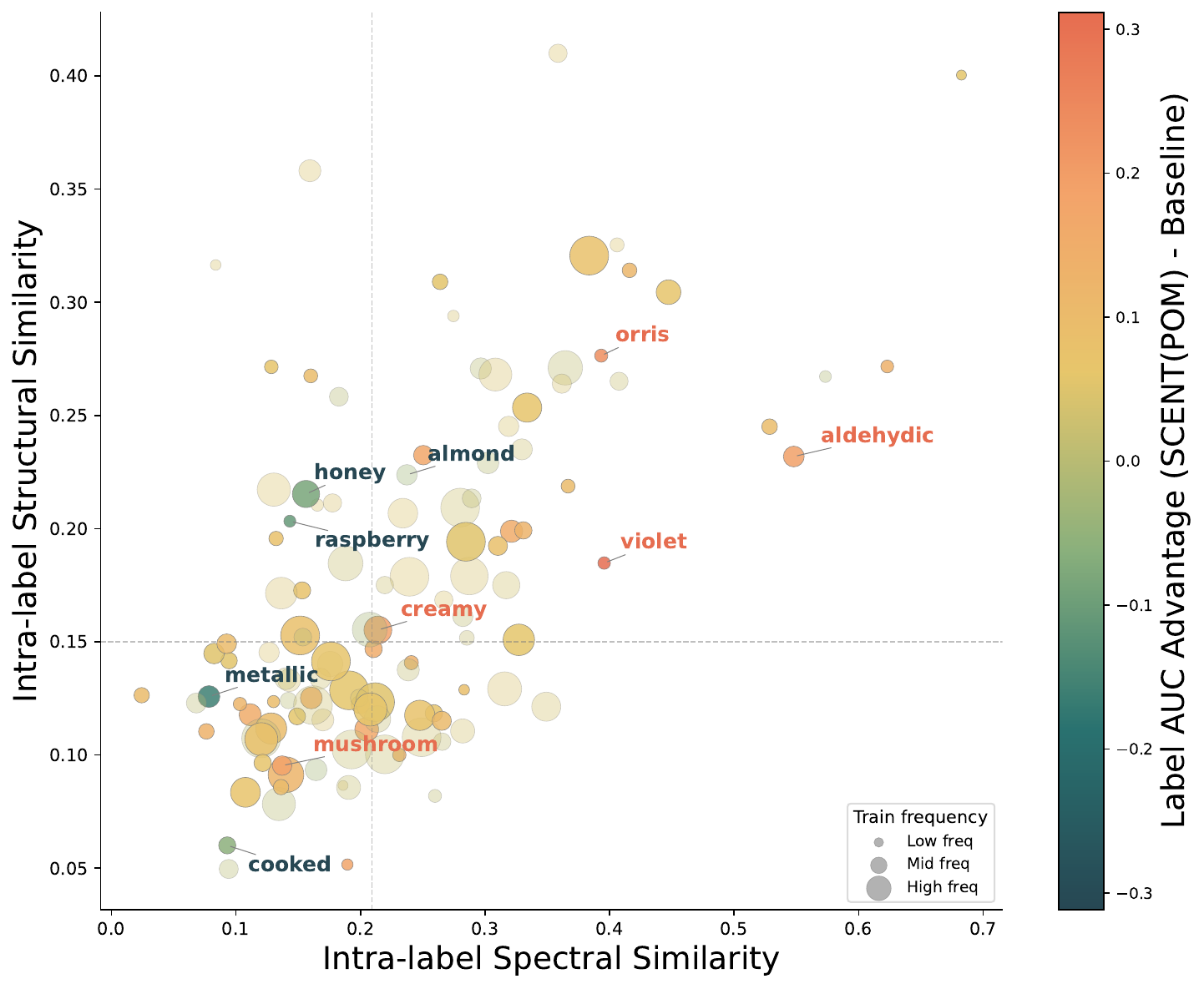}
        \caption{SCENT (Open-POM) vs. EIMS2Vec}
        \label{fig:similarity_pom}
    \end{subfigure}
    \caption{\textbf{Model advantage on test set vs.\ intra-label similarity.} Each point represents one odor descriptor. Axes show mean pairwise intra-label spectral and structural similarity among training molecules. Color encodes performance advantage and point size reflects training frequency.}
    \label{fig:similarity}
\end{figure}

To better understand what drives SCENT's advantage over the unaligned baseline, we compute the mean pairwise intra-label spectral similarity and structural similarity among training molecules for each odor descriptor, and plot them against the per-label AUC difference, as shown in Figure \ref{fig:similarity}. We observe that SCENT outperforms the baseline across the majority of descriptors regardless of their position in the similarity space, confirming that the alignment benefit is general. Point size shows label frequency in training set, and the absence of a clear relationship between bubble size and color confirms that the observed gains are not simply a consequence of label frequency — SCENT's advantage persists across both rare and common descriptors. That is meet with what we have concluded in previous section. 

\subsection{Data scaling}
\label{app:data_scaling}
\begin{figure}[h] %
    \centering
    \includegraphics[width=0.8\linewidth]{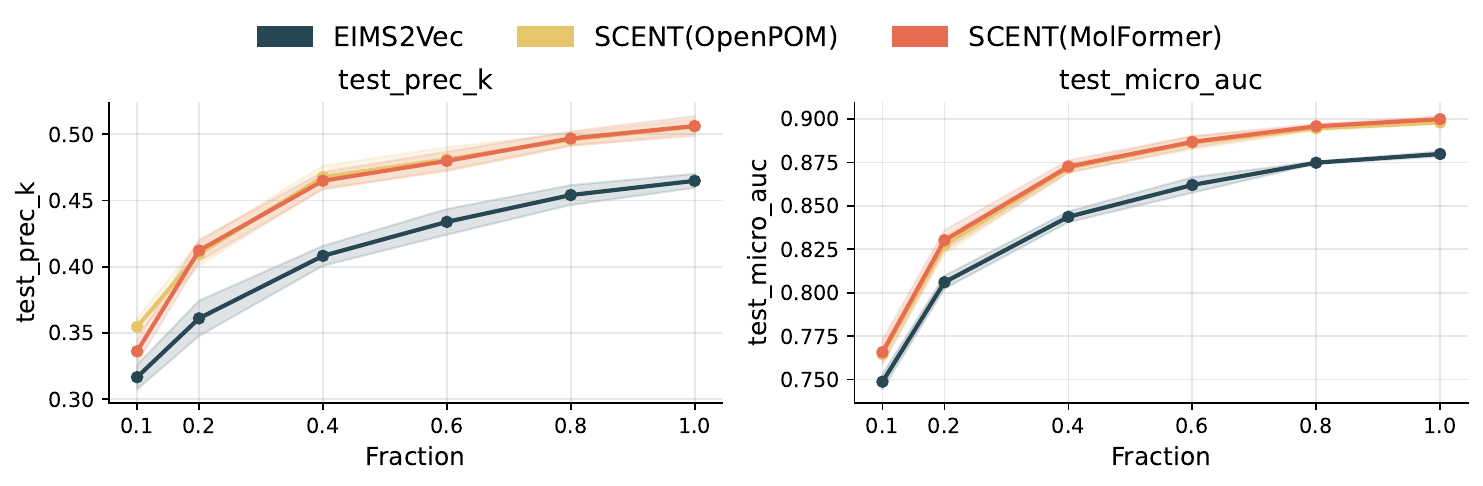}
    \caption{\textbf{Test performance vs.\ training set fraction.} We evaluate SCENT(OpenPOM), SCENT(MolFormer), and EIMS2Vec across varying fractions of the training set (0.1 to 1.0), reporting mean $\pm$ std over 5 folds. SCENT consistently outperforms EIMS2Vec across all fractions and metrics.}
    \label{fig:test_vs_fraction_2} 
\end{figure}

Figure~\ref{fig:test_vs_fraction_2} complements Figure~\ref{fig:test_vs_fraction_1} by reporting test loss and micro-AUC across training set fractions. Both metrics confirm that SCENT maintains a consistent advantage over EIMS2Vec at all data scales, with the gap being most pronounced at small fractions (10–20\%), further supporting the conclusion that alignment pretraining transfers chemical priors that cannot be recovered from spectral signal alone.

\subsection{PCA visualization for full GS-LF dataset}
\begin{figure}[htbp]
    \centering
    \renewcommand\thesubfigure{\Alph{subfigure}} 
    \begin{subfigure}[b]{0.32\linewidth}
        \centering
        \includegraphics[width=\linewidth]{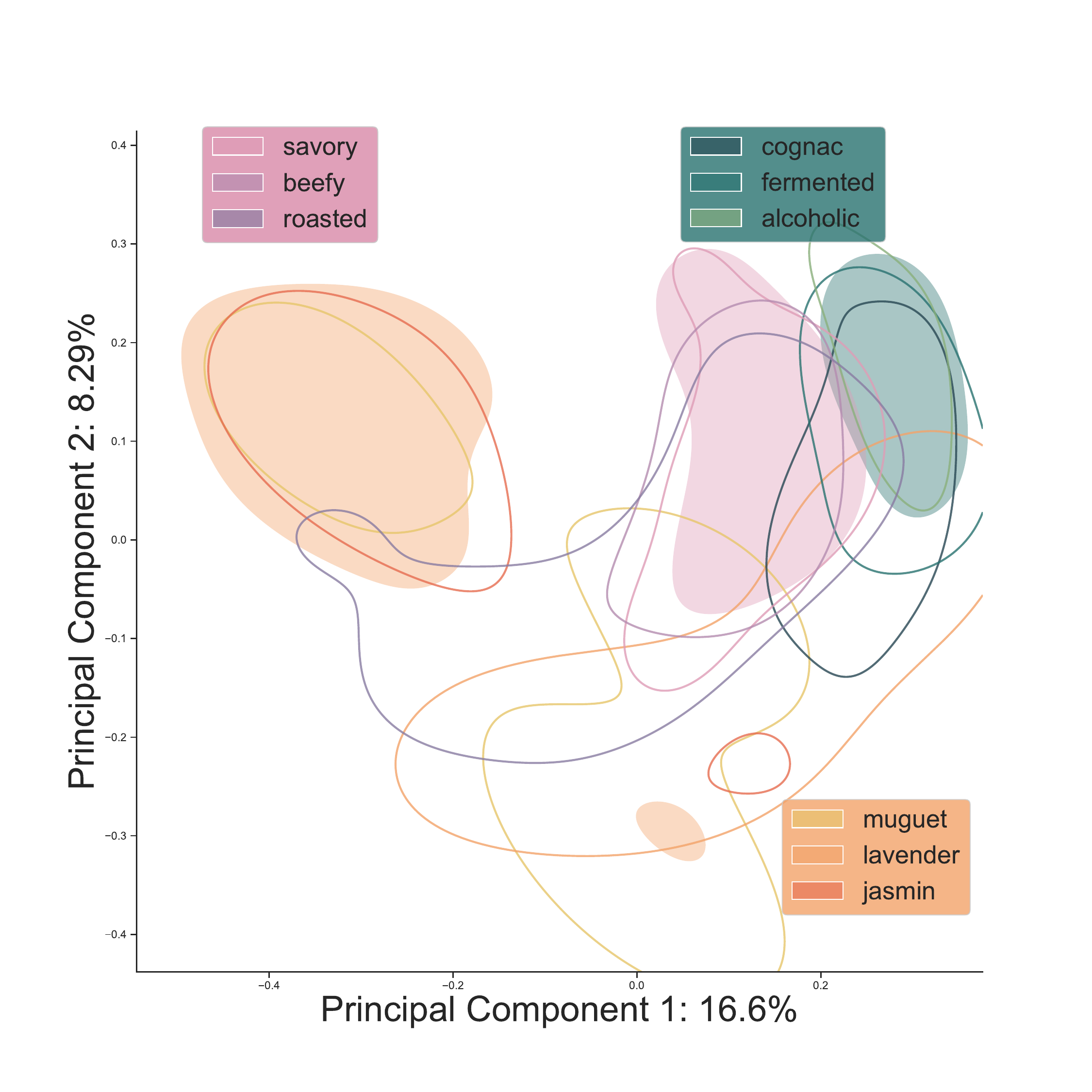} 
        \caption{EIMS2Vec~\citep{10822853}} 
        \label{fig:pca_a}
    \end{subfigure}
    \hfill 
    \begin{subfigure}[b]{0.32\linewidth}
        \centering
        \includegraphics[width=\linewidth]{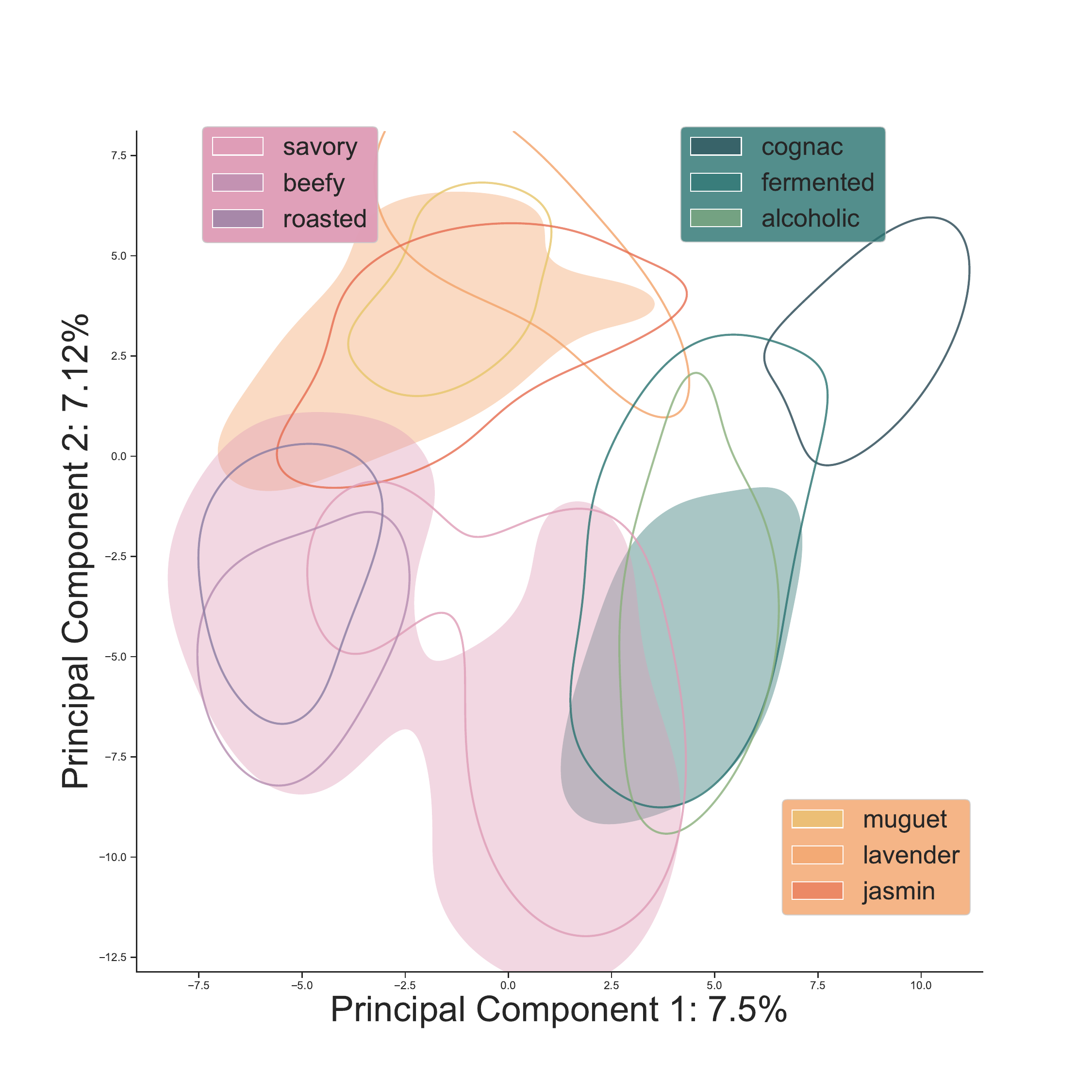}
        \caption{SCENT (MolFormer)}
        \label{fig:pca_b}
    \end{subfigure}
    \hfill
    \begin{subfigure}[b]{0.32\linewidth}
        \centering
        \includegraphics[width=\linewidth]{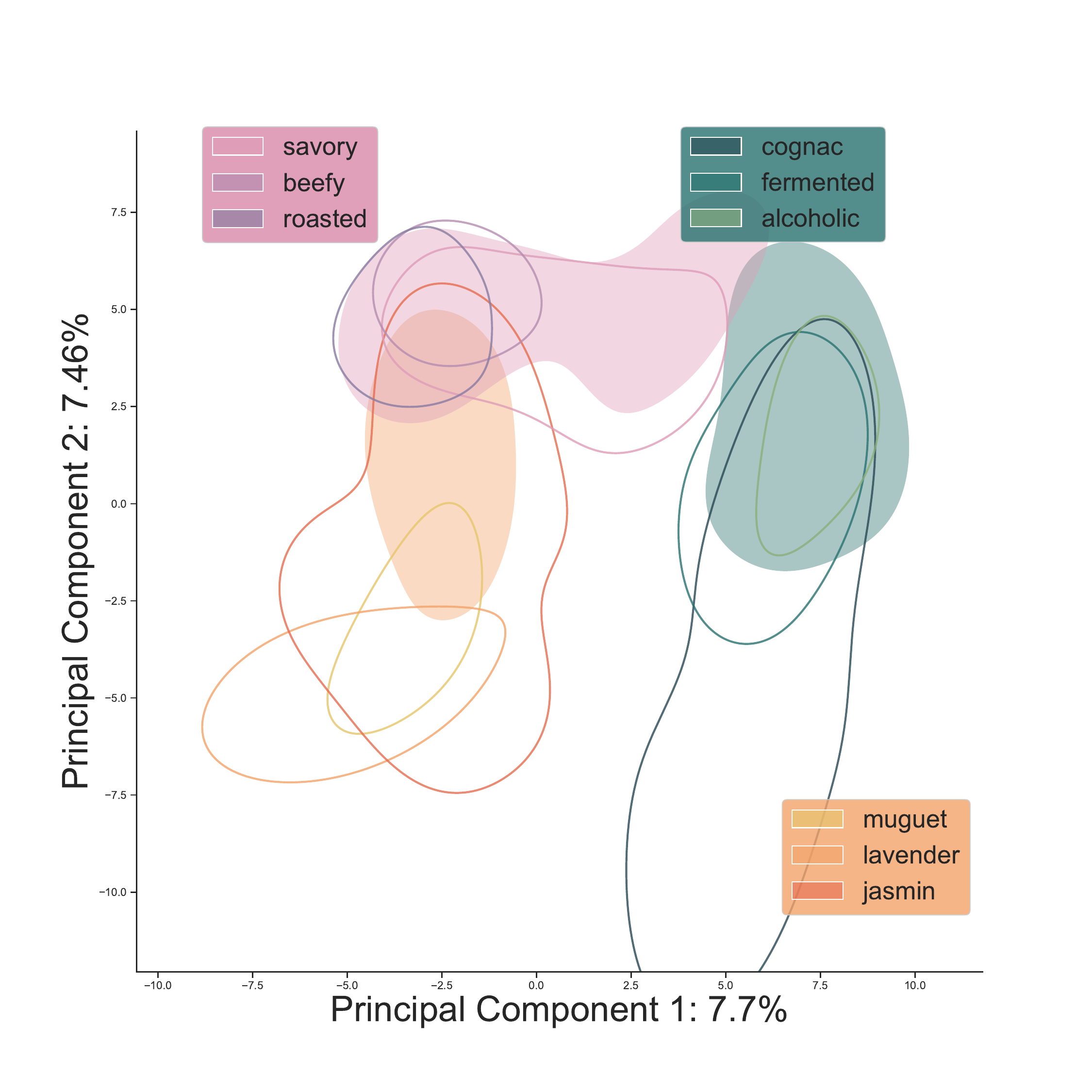}
        \caption{SCENT (Open-POM)}
        \label{fig:pca_c}
    \end{subfigure}
    \caption{\textbf{Visualization of the latent space of different MS encoder models.} The shade highlights the space associated with coarse-grained descriptors, while the contour lines mark regions with more precise descriptors. We employ PCA to reduce the dimensionality of the corresponding embeddings.}
    \label{fig:PCA_full} 
\end{figure}

We visualize the low-dimensional representations of the MS embedding using PCA for different odorants and models. In the aligned space, the embedding spaces of both SCENT variants show markedly improved separation
between odor categories: the ethereal cluster (cognac, fermented, alcoholic) and the meaty cluster (savory, roasted) become more spatially distinct, with reduced overlap between their respective areas. At the same time, we also observe that there is a decrease in the explained variance ratio of the first two principal components (PCs) after alignment (compared to Figure \ref{fig:pca_a}). This result prove that alignment redistributes information across the embedding dimensions, encoding richer odor-relevant structure rather than concentrating variance along a small number of dominant axes.

\section{Additional results: human rating regression}
\label{app:add_result_2}
\subsection{Statistical test results of Pearson $r$}
\label{app:human_reg:stats_tests}
To better understand the statistical significance of the perceptual regression results, we apply the Wilcoxon signed-rank test~\citep{wilcoxon1992individual}, a non-parametric paired test, to each label's Pearson's $r$ vector ($n=21$) in 100 folds cross-validation. The non-parametric nature of the test makes it appropriate here given the small sample size and the absence of a normality assumption on the distribution of per-attribute correlations.
\begin{figure}[t]
    \centering
    \includegraphics[width=1.0\linewidth]{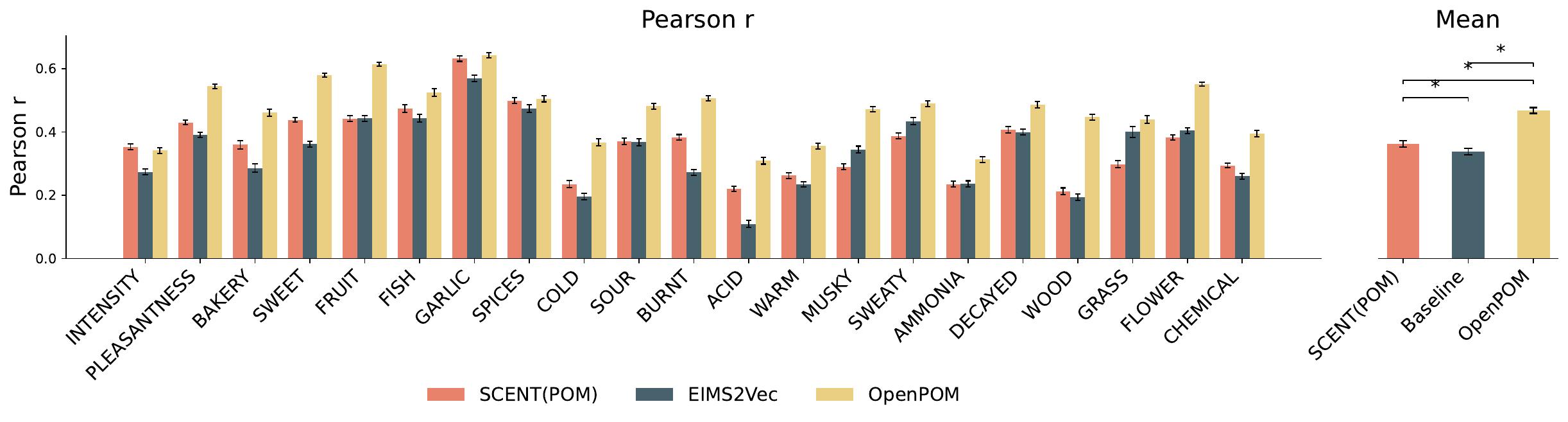}
    \caption{\textbf{Pearson correlation between predicted and human-rated perceptual scores on the DREAM dataset (Open-POM family).} Ridge regression is fitted on frozen embeddings to predict 21 continuous perceptual descriptors (mean $\pm$ std across 100 folds). Statistical significance is assessed via the Wilcoxon signed-rank test~\citep{wilcoxon1992individual} (* $p < .05$).}
    \label{fig:keller_r_pom}
\end{figure}

As shown in Table~\ref{tab:wilcoxon}, both two variants of SCENT significantly outperform EIMS2Vec baseline, suggests that the alignment introduces the structure-related odor information from both supervised and self-supervised chemical model. At the same time, SCENT (MolFormer) also achieves a performance comparable to that of MolFormer, while SCENT (Open-POM) still remains a gap with its chemical teacher. That is expected given that Open-POM embeddings are directly optimized with odor supervision and therefore contain task-specific perceptual structure that cannot be fully recovered from MS signals alone.

\begin{table}[h]
\centering
\caption{Pairwise Wilcoxon signed-rank tests on per-attribute Pearson $r$ (n=21).}
\vspace{1ex}
\label{tab:wilcoxon}
\begin{tabular}{llccc}
\toprule
\textbf{Group} & \textbf{Comparison} & \textbf{W} & \textbf{\textit{p}-value} & \textbf{Sig.} \\
\midrule
\multirow{3}{*}{Open-POM-based}
  & SCENT (Open-POM) vs.\ EIMS2Vec  & 55.0 & 0.035          & *  \\
  & SCENT (Open-POM) vs.\ OpenPOM   & 3.0  & $<$0.001       & *  \\
  & EIMS2Vec vs.\ OpenPOM     & 0.0  & $<$0.001       & *  \\
\midrule
\multirow{3}{*}{MolFormer-based}
  & SCENT (MolFormer) vs.\ EIMS2Vec  & 29.0 & 0.002          & *  \\
  & SCENT (MolFormer) vs.\ MolFormer & 115.0 & 1.000         & —  \\
  & EIMS2Vec vs.\ MolFormer   & 46.0 & 0.014          & *  \\
\bottomrule
\end{tabular}
\end{table}

\section{Additional results: case study}
\label{app:add_result_3}
\subsection{Real-world dataset}
\label{app:case_study:dataset}
We provide the molecules used for the real-world test set in Table~\ref{tab:val_molecules}. The first 28 molecule labels are from the GS-LF dataset. The labels for the last two molecules are from \cite{gs}. All those 30 molecules will not appear in any stage of training in real-world adaption stage to avoid data leakage.

\begin{table}[htbp]
\centering
\small
\caption{Molecules used for real-world perceptual validation. SMILES are in canonical non-stereo form translated by RDKit.}
\vspace{1ex}
\resizebox{\columnwidth}{!}{%
\label{tab:val_molecules}
\begin{tabular}{lccc}
\toprule
\textbf{Canonical SMILES} & \textbf{Formula} & \textbf{Descriptor Number} & \textbf{Odor Descriptors (OD)} \\
\midrule
CC(=O)O                   & C$_2$H$_4$O$_2$        & 3  & pungent, sharp, sour \\
CC(C)=O                   & C$_3$H$_6$O            & 4  & apple, ethereal, pear, solvent \\
O=Cc1ccccc1               & C$_7$H$_6$O            & 9  & almond, bitter, cherry, fruity, nutty, oily, powdery, sharp, sweet \\
CCCC=O                    & C$_4$H$_8$O            & 7  & chocolate, cocoa, fruity, green, malty, musty, pungent \\
CCCCO                     & C$_4$H$_{10}$O         & 7  & alcoholic, balsamic, fermented, oily, solvent, sweet, winey \\
CC(=O)C(C)=O              & C$_4$H$_6$O$_2$        & 5  & buttery, caramellic, creamy, pungent, sweet \\
CC12CCC(CC1)C(C)(C)O2     & C$_{10}$H$_{18}$O      & 5  & camphoreous, cooling, fresh, herbal, medicinal \\
CCC(C)=O                  & C$_4$H$_8$O            & 6  & camphoreous, ethereal, fruity, ketonic, solvent, sweet \\
CC1=CCC2CC1C2(C)C         & C$_{10}$H$_{16}$       & 13 & aromatic, camphoreous, cooling, earthy, fresh, herbal, mint, pine, sharp, sweet, terpenic, warm, woody \\
COc1ccccc1                & C$_7$H$_8$O            & 4  & anisic, ethereal, phenolic, sweet \\
CCCC(=O)OCC               & C$_6$H$_{12}$O$_2$     & 8  & banana, buttery, cognac, ethereal, fruity, juicy, pineapple, ripe \\
CCCC(C)=O                 & C$_5$H$_{10}$O         & 10 & banana, dairy, ethereal, fermented, fruity, ketonic, rummy, sweet, winey, woody \\
C=CC(C)(C)O               & C$_5$H$_{10}$O         & 5  & earthy, ethereal, fruity, herbal, oily \\
CCCCCCOC(C)=O             & C$_8$H$_{16}$O$_2$     & 8  & apple, banana, fatty, fresh, fruity, green, pear, sweet \\
C=CCOC(=O)Cc1ccccc1       & C$_{11}$H$_{12}$O$_2$  & 5  & fruity, honey, pineapple, rummy, sweet \\
C=CC(O)CCCCC              & C$_8$H$_{16}$O         & 12 & earthy, fruity, grassy, green, herbal, melon, mushroom, musty, oily, sweet, vegetable, violet \\
Cc1nccnc1C                & C$_6$H$_8$N$_2$        & 10 & caramellic, cocoa, coffee, green, meaty, musty, nutty, potato, powdery, roasted \\
C=C(C)C1CC=C(C)CC1        & C$_{10}$H$_{16}$       & 11 & aromatic, citrus, fresh, herbal, lemon, orange, phenolic, pine, sweet, terpenic, woody \\
CC(=O)c1cnccn1            & C$_6$H$_6$N$_2$O       & 8  & chocolate, coffee, hazelnut, musty, nutty, popcorn, potato, roasted \\
CCC=CCCO                  & C$_6$H$_{12}$O         & 15 & apple, cortex, earthy, fatty, floral, fresh, fruity, grassy, green, herbal, leafy, melon, oily, pungent, vegetable \\
Cc1ccccc1                 & C$_7$H$_8$             & 1  & sweet \\
O=Cc1ccco1                & C$_5$H$_4$O$_2$        & 6  & almond, caramellic, phenolic, spicy, sweet, woody \\
CCOC(C)=O                 & C$_4$H$_8$O$_2$        & 10 & brandy, ethereal, fruity, grape, green, rummy, sharp, sweet, weedy, winey \\
C=CC(C)=CCC=C(C)C         & C$_{10}$H$_{16}$       & 12 & anisic, citrus, floral, green, herbal, lemon, sweet, terpenic, tropical, vegetable, warm, woody \\
CC1=CCC(=C(C)C)CC1        & C$_{10}$H$_{16}$       & 9  & anisic, citrus, fresh, fruit skin, herbal, lemon, pine, sweet, woody \\
C=CC(C)=O                 & C$_4$H$_6$O            & 5  & ethereal, gassy, pungent, sharp, sweet \\
CC(C)O                    & C$_3$H$_8$O            & 3  & alcoholic, musty, woody \\
Cc1ccccc1O                & C$_7$H$_8$O            & 5  & herbal, leathery, medicinal, musty, phenolic \\
CC=Cc1ccc(OC)cc1          & C$_{10}$H$_{12}$O      & 2  & medicinal, sweet \\
CCCC=CC=O                 & C$_6$H$_{10}$O         & 5  & fresh, fruity, green, leafy, vegetable \\
\bottomrule
\end{tabular}
}
\end{table}

\subsection{Downstream performance when m/z within [50, 180]}
In order to meet the real-world experiment setting, we further retrain the model with m/z only in 50-180 range and report the model performance and report in the Table~\ref{tab:50180_results}. SCENT still remains a stable performance under the restricted m/z range, whereas the unaligned baseline suffers a notably larger performance drop compared to its full-range counterparts. This asymmetry suggests that the contrastive alignment with chemical structure provides representations that are more robust to spectral truncation, makes the learned representations more resilient to the domain shift introduced by real-world measurement constraints.

\begin{table}[t]
\centering
\caption{\label{tab:50180_results}Model performance on filtered GS-LF dataset with MS input within [50, 180]. $\Delta_{\text{full}}$ denotes change relative to full-range MS input.}
\vspace{1ex}
\resizebox{\columnwidth}{!}{%
\begin{tabular}{lccccccc}
\toprule
\textbf{Model} & \textbf{Input} &
\textbf{Micro-AUC}$\uparrow$ & $\Delta_{\text{full}}$ &
\textbf{Weighted-AUC}$\uparrow$ & $\Delta_{\text{full}}$ &
\textbf{Adj.Precision@k}$\uparrow$ & $\Delta_{\text{full}}$ \\
\midrule
EIMS2Vec
& MS[50,180]
& $85.38\pm0.24$ & \loss{2.60}
& $78.06\pm0.49$ & \loss{4.23}
& $41.86\pm0.70$ & \loss{4.63} \\
\midrule
\textbf{SCENT (Open-POM)}
& MS[50,180]
& $88.06\pm0.13^*$ & \loss{1.74}
& $82.73\pm0.32^*$ & \loss{2.72}
& $48.32\pm0.48^*$ & \loss{2.28} \\
\textbf{SCENT (MolFormer)}
& MS[50,180]
& $87.99\pm0.32^*$ & \loss{2.00}
& $82.44\pm0.53^*$ & \loss{3.27}
& $46.56\pm0.56^*$ & \loss{4.06} \\
\bottomrule
\end{tabular}
}
\vspace{-2ex}
\end{table}

\section{Label distribution}
\label{app:label}
\begin{figure}[htbp]
    \centering
        \centering
        \includegraphics[width=\textwidth]{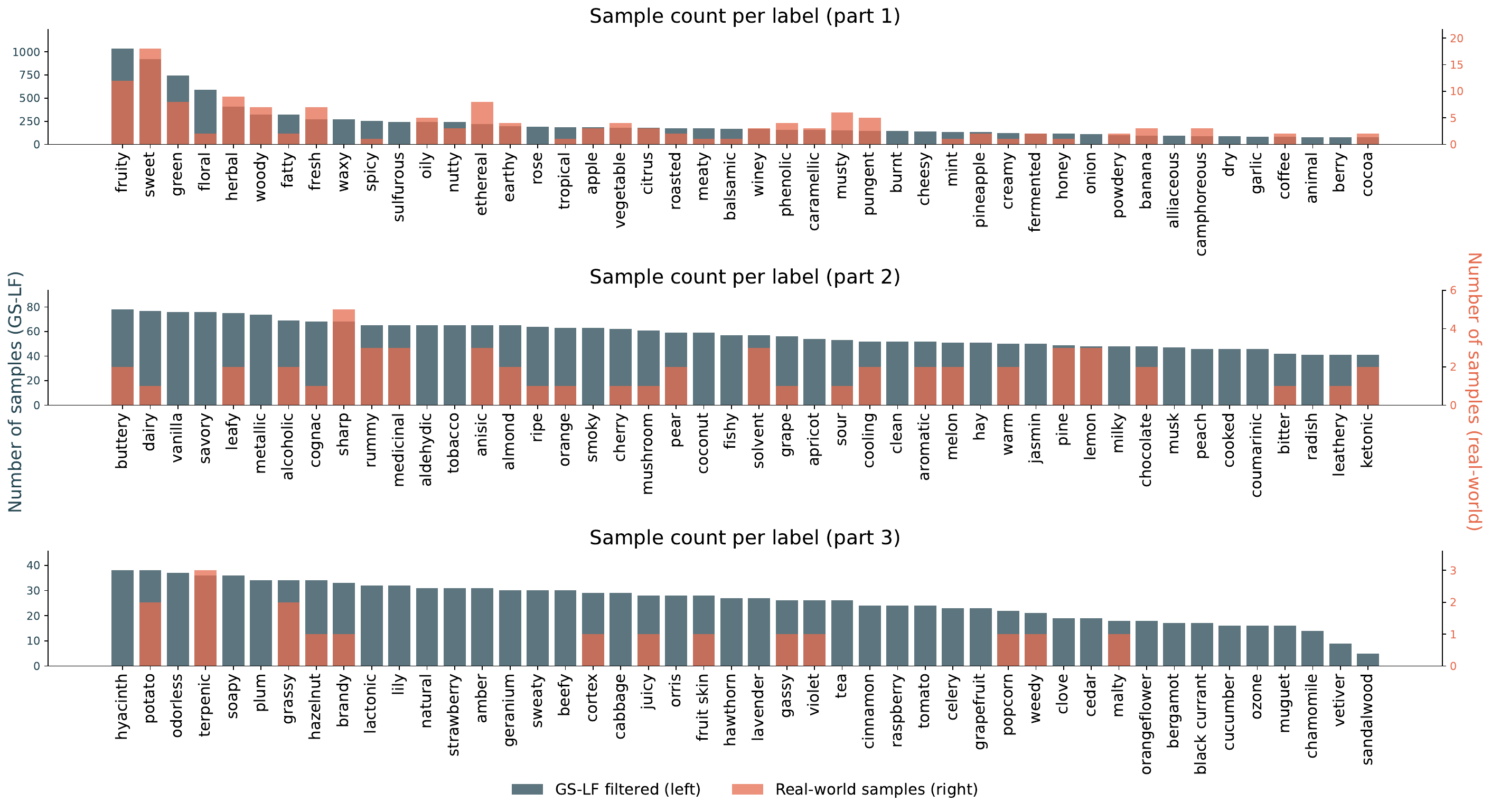}
        \caption{\textbf{Label distribution among filtered GS-LF dataset and real sample dataset.} }
        \label{fig:label_dist}
\end{figure}

GS-LF dataset has a long-tail feature. To better understand how label distribution affects the results, and to what extend the real-world dataset label distribution similar with the training dataset, we visualize the label distribution among the full GS-LF dataset and real sample validation set. According to Figure ~\ref{fig:label_dist}, both datasets existing long-tail distribution. The real-world measurements not only covers on the common odors, but also contains the labels with less samples in the GS-LF dataset, makes the results more credible.

\begin{figure}
    \centering
    \begin{subfigure}{0.48\linewidth}
        \includegraphics[width=\linewidth]{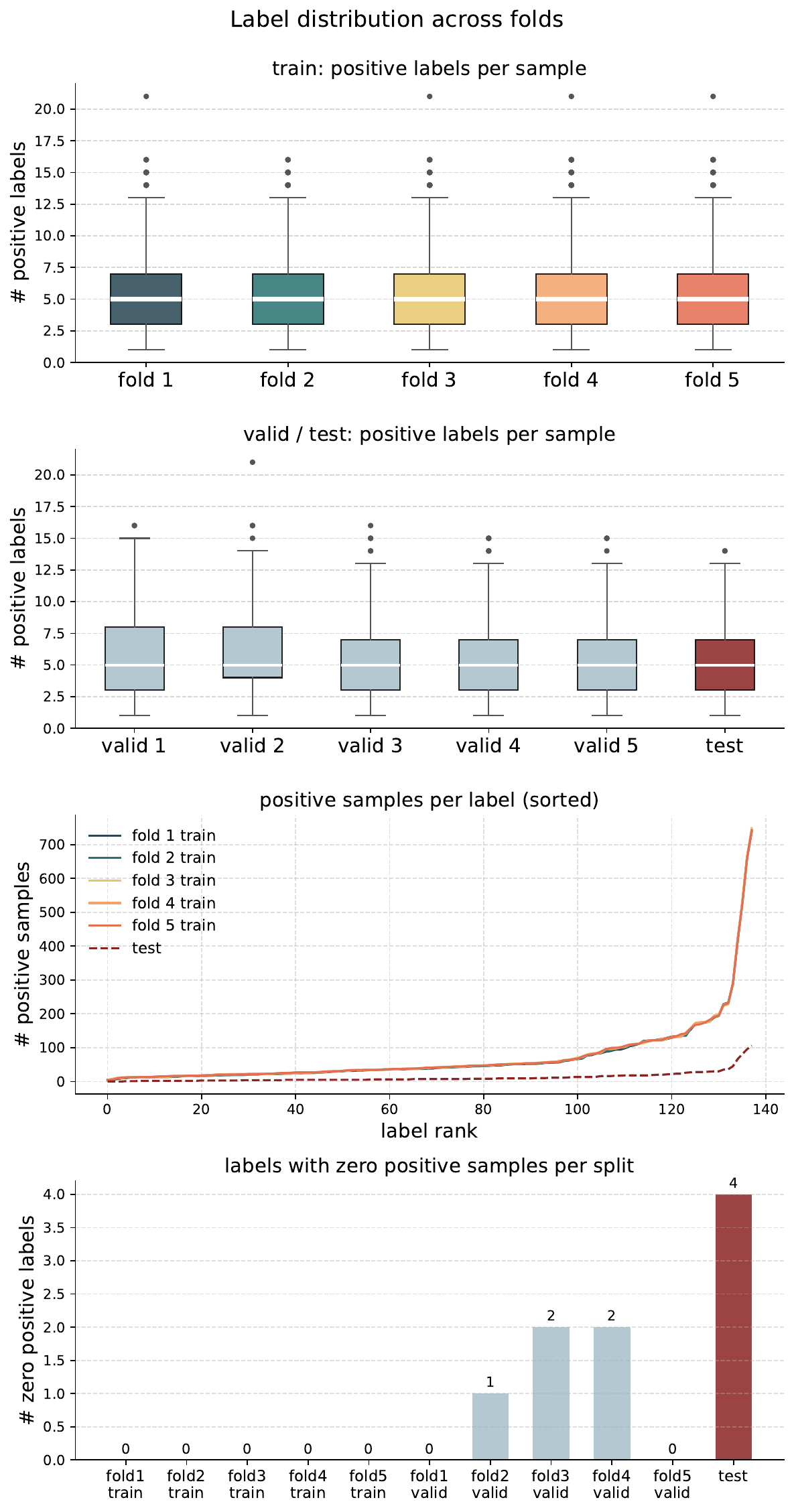}
        \caption{\textbf{Full spectrum}.}
        \label{fig:label_distribution_full}
    \end{subfigure}
    \hfill
    \begin{subfigure}{0.48\linewidth}
        \includegraphics[width=\linewidth]{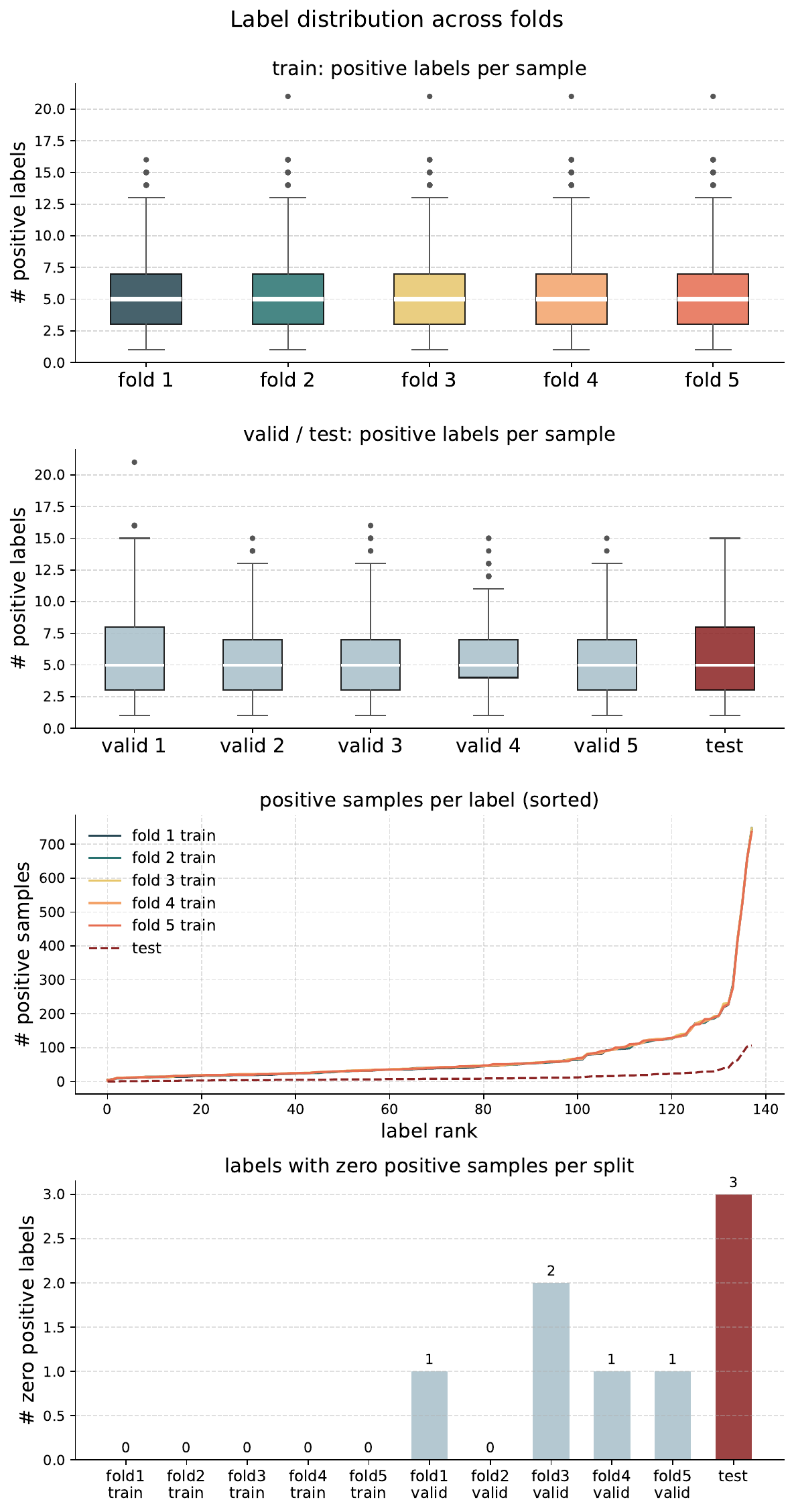}
        \caption{\textbf{m/z in [50, 180] downstream task}.}
        \label{fig:label_distribution_50180}
    \end{subfigure}
    
    \caption{Label distributions of the downstream task.}
    \label{fig:label_distributions}
\end{figure}

For each fold and test set for odor prediction task, the label distribution for full spectra experiment and [50, 180] experiment are shown in Figure \ref{fig:label_distribution_full} and Figure \ref{fig:label_distribution_50180} respectively. The top panels show the distribution of positive labels per sample across folds. Both training and validation splits maintain a consistent median of approximately 5 positive labels per sample, with the test set showing a similar range, confirming that iterative stratified splitting preserves the label co-occurrence structure across all splits. The bottom-left panel shows the number of positive samples per label sorted by frequency, reflecting the long-tail nature of the GS-LF dataset. The bottom-right panel reports the number of labels with zero positive samples in each split. All training folds contain no zero-positive labels, while a small number appear in validation folds and the test set, which is expected given the class imbalance in rare odor categories. 

\clearpage

\end{document}